\documentclass[10pt,twocolumn,letterpaper]{article}

\usepackage[pagenumbers]{cvpr} 

\usepackage[dvipsnames]{xcolor}
\usepackage{xspace}
\usepackage{makecell}
\usepackage{multirow}
\usepackage{colortbl}
\usepackage{algorithmic}
\usepackage[linesnumbered,ruled,vlined]{algorithm2e}
\usepackage{threeparttable}
\usepackage{bbding}
\usepackage{diagbox}
\usepackage{float}

\newcommand{\VspaceBefore}{\vspace{-3.3mm}}

\newcommand{\vidar}{*\xspace}
\newcommand{\datr}{**\xspace}
\newcommand{\cblank}{-}

\newcommand{\fullsetting}{{a}ctive {t}est-time {a}daptation for {s}emantic {seg}mentation\xspace}
\newcommand{\setting}{ATASeg\xspace}
\newcommand{\method}{ATASeg\xspace}
\newcommand{\parastart}[1]{\noindent \textbf{#1}}

\newcommand{\vspaceBeforCaption}{\vspace{-2mm}}
\newcommand{\vspaceAfterCaption}{\vspace{-3mm}}

\definecolor{Gray}{gray}{0.9}
\newcommand{\rowgray}{\rowcolor{Gray}}
\newcommand{\fig}{Fig.}
\newcommand{\tab}{Tab.}

\newcommand{\dis}{\mathcal{P}}
\newcommand{\imgseg}{\mathbf{I}}
\newcommand{\labseg}{\mathbf{Y}}
\newcommand{\actlab}{\widetilde{\mathbf{Y}}}
\newcommand{\pre}{\mathbf{P}}
\newcommand{\varA}{\mathcal{A}}
\newcommand{\varL}{\mathcal{L}}
\newcommand{\Bzero}{ATASeg-B0\xspace}
\newcommand{\Bone}{ATASeg-B1\xspace}
\newcommand{\Real}{\mathbb{R}}

\definecolor{cvprblue}{rgb}{0.21,0.49,0.74}
\usepackage[pagebackref,breaklinks,colorlinks,citecolor=cvprblue]{hyperref}

\title{Few Clicks Suffice: Active Test-Time Adaptation for Semantic Segmentation}

\author{Longhui Yuan \quad Shuang Li \quad Zhuo He \quad Binhui Xie \\
Beijing Institute of Technology, Beijing, China \\
{\tt\small \{longhuiyuan,shuangli,zhuohe,binhuixie\}@bit.edu.cn}}

\begin{document}
\maketitle
\begin{abstract}
    Test-time adaptation (TTA) adapts the pre-trained models during inference using unlabeled test data and has received a lot of research attention due to its potential practical value. 
    Unfortunately, without any label supervision, existing TTA methods rely heavily on heuristic or empirical studies.
    Where to update the model always falls into suboptimal or brings more computational resource consumption.
    Meanwhile, there is still a significant performance gap between the TTA approaches and their supervised counterparts.
    Motivated by active learning, in this work, we propose the \fullsetting setup.
    Specifically, we introduce the human-in-the-loop pattern during the testing phase, which queries very few labels to facilitate predictions and model updates in an online manner. 
    To do so, we propose a simple but effective \method framework, which consists of two parts, \ie, {\it model adapter} and {label annotator}. 
    Extensive experiments demonstrate that \method bridges the performance gap between TTA methods and their supervised counterparts with only extremely few annotations, even one click for labeling surpasses known SOTA TTA methods by $2.6\%$ average mIoU on ACDC benchmark.
    Empirical results imply that progress in either the model adapter or the label annotator will bring improvements to the \method framework, giving it large research and reality potential.
\end{abstract}


\section{Introduction}
\label{sec:intro}
\begin{figure}
    \centering
    \includegraphics[width=0.99\linewidth]{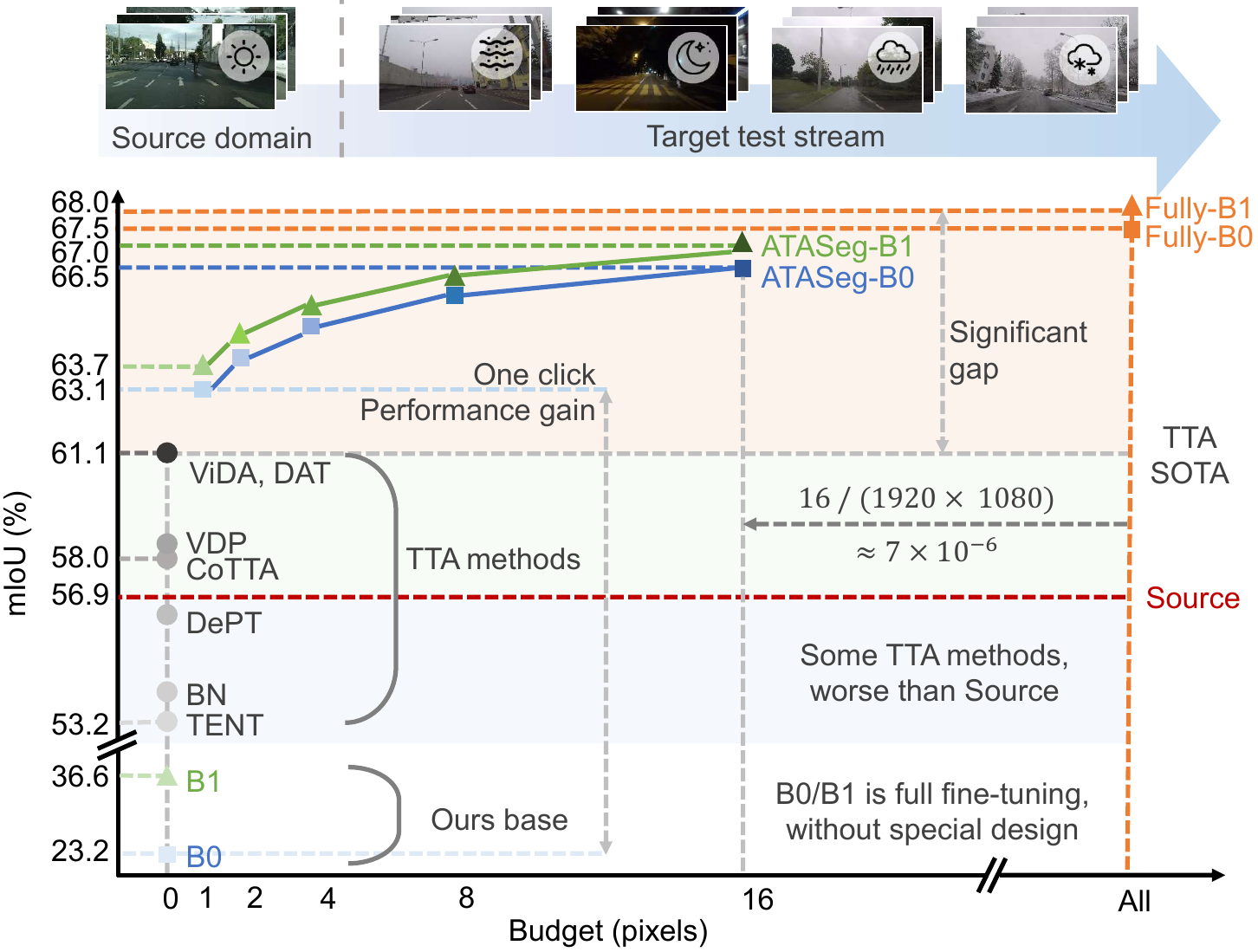}
    \vspaceBeforCaption
    \caption{{\bf Performance} vs. {\bf budget} on  Cityscapes$\to$ACDC of existing test-time adaptation approaches ($\bullet$)~\cite{tent_wang2020,BN_Stat, DePT,cotta,VDP,ViDA,DAT}, our proposed \method-B0 ($\blacksquare$) and \method-B1 ($\blacktriangle$), under the CTTA~\cite{cotta} test protocol. As demonstrated, despite being carefully designed, DePT~\cite{DePT} still performs worse than the source model (Source). Meanwhile, with intuitively selected trainable parameters, BN~\cite{BN_Stat} and TENT~\cite{tent_wang2020} achieve excellent performance in classification but show ineffectiveness in segmentation. Furthermore, a significant performance gap persists between SOTA methods, \ie, DAT~\cite{DAT} and ViDA~\cite{ViDA}, and supervised counterparts. Surprisingly, with only one click for annotating, \method surpasses DAT and ViDA by a large margin. Finally, with extremely few annotations, \ie, $1\sim16$ pixels, \method establishes bridges between TTA methods and their supervised counterparts.}
    \label{fig:introduction}
    \vspaceAfterCaption
    \vspace{-3mm}
\end{figure}

Semantic segmentation, crucial for pixel-level image comprehension, has applications in various vision-based domains, such as autonomous driving~\cite{TeichmannWZCU18, YangYZLY18}, robot manipulation~\cite{SheikhMLSBS20, ValadaVDB17}, and medical diagnostics~\cite{ronneberger2015u-net, TaghanakiACCH21}. 
In recent years, deep neural networks (DNNs) have demonstrated remarkable achievements in the field of semantic segmentation~\cite{chen2018deeplabv3plus, XieWYAAL21SegFormer,StrudelPLS21}. 
However, their performance significantly degrades in the presence of the distribution shift~\cite{dataset_shift_in_ML09}, where the training data and test data originate from different data distributions.
Many efforts have been dedicated to addressing this issue, among which test-time adaptation shows substantial promise for real-world applications~\cite{tent_wang2020, HuSGLCCZZ21, YiYWLTK23}. 

Test-time adaptation (TTA)~\cite{rotta,grotta,liang2023ttasurvey,Wang23ttasurvey,tent_wang2020,cotta,EATA,ODS} attempts to address the distribution shift during the online inference stage with only access to unlabeled test data, demonstrating considerable potential in scenarios demanding real-time predictions and stringent data privacy.
However, due to the absence of label supervision, adaptation at test time always suffers from error accumulation.
Most existing methods heavily depend on empirical or heuristic studies to address this problem, especially in terms of which parts of parameters to update. 
According to that, previous literature generally falls into three channels: 
(i) Partial fine-tuning~\cite{tent_wang2020,EATA,Niu00WCZT23}, specifically, where only some parts of the parameters are considered learnable while the remaining are held fixed.
(2) Additional fine-tuning~\cite{DePT,VDP,ViDA,song2023ecotta,DAT}, which introduces extra trainable modules into the original model without altering pre-trained parameters.
(3) Full fine-tuning~\cite{cotta,BrahmaR23}, where all parameters are adapted, always necessitating parameter restoration and multiple data augmentations to guarantee stability.

While these approaches have proven their effectiveness, deciding the number of trainable parameters in them is still reliant on human intuitions.
This creates a dilemma for TTA methods, \ie, having too many parameters involved in adaptation can lead to instability, and needs additional operations to mitigate adverse effects, thereby increasing computing resource consumption, while too few will lead to insufficient model learning and limited generalization ability.
Without any label supervision, it is challenging to solve this problem under the TTA setup.
Meanwhile, as illustrated in \fig~\ref{fig:introduction}, a significant performance gap persists between the SOTA TTA methods and the supervised counterparts. 
Notably, full fine-tuning without special design (B0/B1) suffers severely from error accumulation.
In addition, even some well-designed methods (DePT~\cite{DePT}) perform worse than the source model.
These problems increase risk when applying TTA in scenarios with high-security or high-accuracy requirements, such as autonomous driving and medical image analysis.
A new paradigm is urgently needed to address aforementioned issues while satisfying the data privacy and time efficiency requirements of TTA.

Fortunately, active learning (AL)~\cite{settles2009active,RenXCHLGCW22,OALsurvey} has provided some possible clues for solving this predicament, which is annotating the most informative pixels within a very limited budget.
Nonetheless, employing AL algorithms directly does not meet our expectations.
Firstly, the pool-based AL approaches~\cite{ENT,BvSB_2009_CVPR,Shin0CWPK21,NingL0BYY0021,RIPU} require multiple training-selection rounds and acquire labels for massive pixels, violating the requirement of timeliness.
Secondly, although the deep stream-based AL studies~\cite{WangADSG21,BanZTBH22,SaranYK0A23} select samples and train the models in an online manner, they largely rely on the independent and identically distributed (i.i.d.) assumption. 
However, the test distribution may change over time due to factors such as alterations in the surrounding environment~\cite{cotta,rotta}, thus breaking the i.i.d. prerequisite of stream-based AL. 
Furthermore, stream-based research in the field of image semantic segmentation is still left blank.

Inspired by the above analysis, we propose the \fullsetting setup.
More specifically, we introduce the human-in-the-loop pattern during the test phase, which queries very few labels for the most informative pixels to enhance segmentation and model adaptation in an online manner.
Simultaneously, to ensure practicality, different from conventional AL studies, only extremely few annotations are allowed, \ie, the active budget never exceeds 16 pixels.
Surprisingly, with the inclusion of label information, albeit very limited, the error accumulation and the mentioned dilemma of determining trainable parameters faced by TTA are resolved. 
As illustrated in \fig~\ref{fig:introduction}, only one-click annotating brings astonishing improvements to full fine-tuning methods (B0/B1) without special designs, even surpassing the SOTA TTA methods directly by $2.0\%$ and $2.6\%$ respectively, indicating extraordinary potential of the proposed setting.

To pave a new path, this work introduces a simple but effective \method framework, which consists of the model adapter and label annotator. 
For the model adapter, we develop two simple full fine-tuning approaches, which are denoted as B0 and B1 respectively. 
Meanwhile, four active selection strategies are adopted for the label annotator, including random~(Rand), softmax entropy~(Ent)~\cite{entropy_2014_IJCNN}, region impurity and prediction uncertainty~(RIPU)~\cite{RIPU} and best vs. second best~(BvSB)~\cite{BvSB_2009_CVPR}.
With extensive experiments, we demonstrate the remarkable performance and excellent scalability of \method framework.

Contributions of this work can be summarized as:
\begin{itemize}
    \item To our knowledge, we are the first to propose the \fullsetting setup, which allows very few annotations during test time.
    \item A simple but effective \method framework is introduced, which is compatible with any model adapter and label annotator, showing great flexibility.
    \item Two preliminary model adapters cooperating with four distinct label annotators are developed to serve as the strong baselines for the new field.
    \item Extensive experiments on Cityscapes$\to$ACDC and Cityscapes$\to$Cityscapes-C under test protocols of both FTTA~\cite{tent_wang2020} and CTTA~\cite{cotta} validate the effectiveness and scalability of \method framework.
\end{itemize}

\section{Related Work}
\label{sec:related}

\begin{figure*}[!htb]
    \centering
    \includegraphics[width=0.99\linewidth]{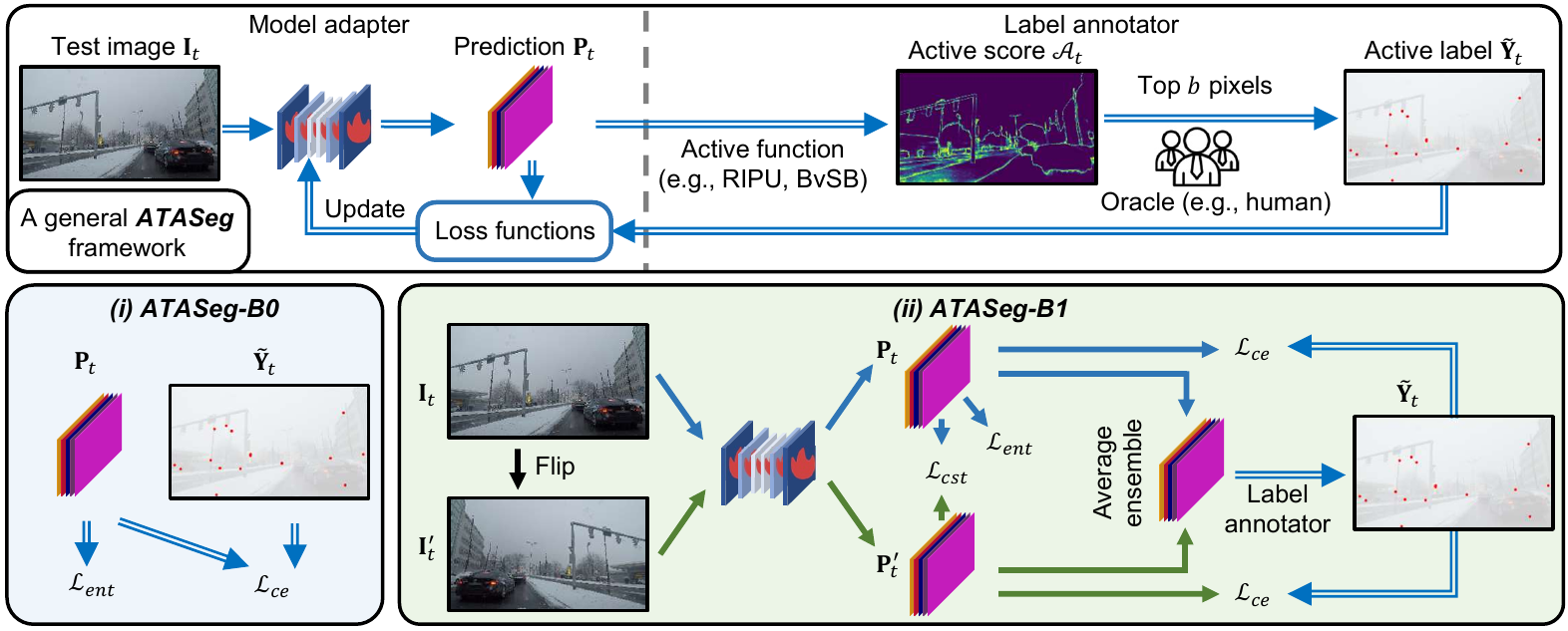}
    \vspaceBeforCaption
    \caption{{\bf The framework of ATASeg.} We first utilize the model adapter to make the pixel-level prediction $\pre_t$ for each test sample $\imgseg_t$. Then the label annotator recieves $\pre_t$ and calculates an active score $\varA_t$ via active functions, further selecting top $b$ pixels to obtain active labels $\actlab_t$ queried from the Oracle. At last, the model adapter updates the segmentation network using $\pre_t$ and $\actlab_t$. Furthermore, we introduce two model adapters \Bzero and \Bone to consolidate the foundation and demonstrate the necessity of our setup.} 
    \label{fig:framework}
    \vspaceAfterCaption
    \vspace{-3mm}
\end{figure*}

\parastart{Active Learning (AL)}
aims to maximize a model's performance while annotating as few samples as possible~\cite{settles2009active,RenXCHLGCW22,OALsurvey,xie2023annotator}, which is roughly categorized into pool-based and stream-based methods.
The pool-based AL is primarily divided into uncertainty sampling~\cite{hanneke2014theory, BvSB_2009_CVPR, ENT}, representative sampling~\cite{CoreSet_2019_ICLR,sinha2019variational} and the hybird of the two sampling~\cite{BADGE_2020_ICLR, zhdanov2019diverse}.
However, these conventional works are inadequate for addressing the frequently occurring domain shift issues.
To this end, Active Domain Adaptation (ADA)~\cite{rai2010domain, AADA_WACV, Fu_2021_CVPR, CULE_2021_ICCV, xie2022active} is proposed, with the objective of selecting target samples for domain adaptation.
Though impressive, these AL algorithms involve multiple training-selection rounds and the acquisition of labels for massive pixels, violating the requirement of timeliness.

The standard stream-based AL methods~\cite{Cesa-BianchiGZ04,ShanALEnsemble,HaoLZZHM18,LuZH16,DasguptaKM09,LoyStream12,ShahM20} are all developed within linear frameworks, which may not be suitable for deep models, because updating deep models is significantly more computationally expensive and challenging. Lately, several deep stream-based AL~\cite{WangADSG21,BanZTBH22,SaranYK0A23} studies are conducted to combine stream-based AL methods with deep neural networks. Nevertheless, they largely rely on the i.i.d. assumption, which may not hold owning to continuous changes in the test distribution. 
Most recently, Yu \etal~\cite{activetest} explore the incorporation of active testing (AT) with AL,
but this topic falls well beyond the scope of this paper.

As to AL for segmentation, most existing work~\cite{CasanovaPRP20, siddiqui_2020_CVPR, CaiXLF21, Wu_2021_ICCV, NingL0BYY0021, Shin0CWPK21, RIPU, xie2023annotator} can be categorized as approach utilizing pool-based AL.
Employing AL algorithms directly
does not meet our expectations.

\parastart{Test-Time Adaptation (TTA)}~\cite{rotta,grotta,liang2023ttasurvey,Wang23ttasurvey,tent_wang2020,cotta,EATA,ODS,Niu00WCZT23,zhang2022memo,note,TTN,RMT,BrahmaR23,song2023ecotta,ChoiYCY22,niid_boudiaf2022parameter,WangZYZL23,zhang2023domainadaptor,ParkTTA2023,NiloyTTA2023,YuLDLZY23} focuses on mitigating distribution shift during the online inference stage with only access to unlabeled test data. 
Based on which parts of parameters to update, previous literature can be generally categorized into three types: 
(1) Partial fine-tuning: These methods update only some parts of the parameters while the remaining are held fixed. 
Some literature~\cite{tent_wang2020,rotta,grotta,note} update the parameters of the model's batch normalization (BN) layers directly, while EATA~\cite{EATA} and TTN~\cite{TTN} estimate the importance of BN layers parameters and limit drastic alterations to significant parameters. 
Different from the aforementioned methods, SAR~\cite{Niu00WCZT23} adapts parameters of batch-agnostic norm layers, \ie, group norm and layer norm.
(2) Additional fine-tuning: Extra trainable modules are introduced into the original model without altering the pre-trained parameters. 
In~\cite{VDP, DePT}, the learnable visual prompts are well designed and fine-tuned in the adaptation phase. 
EcoTTA~\cite{song2023ecotta} introduces meta networks attached to the original networks that are adapted during test time. 
(3) Full fine-tuning: These methods adapt all model parameters~\cite{zhang2022memo, ChoiYCY22, cotta, BrahmaR23, RMT, WangZYZL23}. 
Suffering from instability caused by adapting massive parameters, additional operations needed to be conducted to mitigate adverse effects, like multiple data augmentations and parameter restoration.

Recently, some TTA works~\cite{ViDA,DAT} in image semantic segmentation fall into the partial fine-tuning category.
Nonetheless, 
a substantial performance gap remains between these methods and the supervised counterparts. 
Furthermore, error accumulation and the dilemma caused by reliance on human intuitions persist under TTA setup. 
To this end, we propose the active test-time adaptation for semantic segmentation setup, where only very few labels are queried for the most informative pixels, further enhancing segmentation and model adaptation in an online manner.

\section{Method}
\label{sec:method}

\subsection{Preliminary}
\parastart{Active test-time adaptation for semantic segmentation} aims to adapt a segmentation network $f_{\theta_0}$ pre-trained on the source domain $\mathcal{D}=\{(\imgseg^{\mathcal{S}}, \labseg^{\mathcal{S}})\}$ to an online test stream $\{\imgseg_t | \imgseg_t\sim\dis_{test}\}_{t=0}^{\infty}$ while allowing the model to query very few annotations within the limited active budget $b$.
More specifically, at time step $t$, given a single test image $\imgseg_t$, the model $f_{\theta_t}$ firstly makes predictions $f_{\theta_t}(\imgseg_t)$ for it. Then, after obtaining labels of the most informative pixels, the model $f_{\theta_t}$ will update itself for future inference.
In the context of TTA, a series of works~\cite{cotta, note, niid_boudiaf2022parameter, rotta, grotta} have been derived by improving the assumptions on the test distribution $\dis_{test}$, resulting in a lot of different test protocols.
In this work, we focus on two widely studied test protocols, \ie, FTTA~\cite{tent_wang2020} and CTTA~\cite{cotta}.
FTTA assumes the test distribution is a fixed data distribution, while CTTA considers $\dis_{test}$ changes continually as $\dis_0,\dis_1,\cdots$ as time goes on.

In fact, this setup is largely driven by the core idea of maximizing model performance with minimum labeling cost.
As mentioned in Section~\ref{sec:intro}, without any labeling information, TTA can only rely on empirical or heuristic design to reduce the accumulation of errors, increasing risk in real-world applications.
Distinctively, our goal is to boost performance with extremely few costs during the test-time adaptation process.
Such a trade-off is valuable for scenarios with high-security or high-accuracy requirements like autonomous driving and medical image analysis.

\parastart{\method framework.} 
To fulfill the above objective, we propose a general \method framework.
An overall overview is shown in \fig~\ref{fig:framework}.
More specifically, at time step $t$, the pixel-level prediction $\pre_t$ of test sample $\imgseg_t$ from the model adapter is fed into the label annotator. 
Then the label annotator calculates an active score $\varA_t$ for each pixel via active functions.
Following this, active labels $\actlab_t$ of the top $b$ pixels with the highest active score are queried from the Oracle (e.g., human).
Subsequently, the model adapter updates the segmentation network with the following equation
\begin{small}
    \begin{equation}
        \theta_{t+1} = {\arg\min}_{\theta_t}~\varL_{all}(\pre_t,\actlab_t;\theta_t)\,.
    \end{equation}
\end{small}%
As we can find, \method is compatible with almost all existing TTA methods for model updating and active strategies for label annotating.
To consolidate the foundation and demonstrate the necessity of the new setting, two simple but effective model adapters, \ie, \Bzero and \Bone (also shown in \fig~\ref{fig:framework}), combined with four existing label annotators are introduced to serve as strong baselines.

\subsection{Model Adapter}
Given a test image $\imgseg_t$, the model adapter will infer for it and then update the segmentation network $f_{\theta_t}$ with the cooperation of active labels $\actlab_t$.
In this part, we first introduce a naive solution that directly updates $f_{\theta_t}$ with entropy minimization and cross-entropy loss, denoted as \Bzero.
To further improve the performance, another model adapter that adopts the idea of consistency regularization is developed, called \Bone.
 
\parastart{\Bzero.} 
We first compute the prediction $\pre_t$ of $\imgseg_t$ by $\pre_t = f_{\theta_t} \left( \imgseg_t \right) \in \Real^{H\times W\times C}$,
where $H \text{ and } W$ are height and width of the test image, and $C$ represents the number of categories.
Then the active labels $\actlab_t \in\Real^{H\times W}$ are obtained by feeding $\pre_t$ into the label annotator.
After that, the model $f_{\theta_t}$ is updated by the following loss function
\begin{small}
    \begin{equation}
        \varL_{B0} = \varL_{ce}(\pre_t,\actlab_t) + \lambda_{ent} \varL_{ent}(\pre_t)\,,
        \label{eq:b0}
    \end{equation}
\end{small}%
where $\varL_{ce}$ is the cross-entropy (CE) loss expressed as 
\begin{small}
    \begin{equation}
        \varL_{ce}(\pre_t,\actlab_t) = -\frac{1}{|\actlab_t|}\sum_{(i,j)\in\actlab_t} \sum_{c=1}^{C}\mathbb{I}_{\{\actlab_t^{(i,j)}=c\}} \log \pre_t^{(i,j,c)}\,,
    \end{equation}
\end{small}%
$\varL_{ent}$ is the entropy of the prediction, calculated as
\begin{small}
    \begin{equation}
        \varL_{ent}(\pre_t) = -\frac{1}{|\imgseg_t|}\sum_{(i,j)\in\imgseg_t} \sum_{c=1}^{C} \pre_t^{(i,j,c)}  \log \pre_t^{(i,j,c)}\,,
    \end{equation}
\end{small}%
and $\lambda_{ent}$ is the weight of entropy minimization.

\parastart{\Bone} further considers consistency learning.
Firstly, we employ horizontal flip augmentation on the original test image $\imgseg_t$, creating a weakly augmented view $\imgseg_t'$.
After that the prediction $\overline{\pre}_t$ is given by the average ensemble on predictions of original and augmented view, computed by $\pre_t = f_{\theta_t} \left( \imgseg_t \right),\pre_t' = f_{\theta_t} \left( \imgseg_t' \right),\overline{\pre}_t=\frac{1}{2}\left(\pre_t + \pre_t'\right)$.
Then the label annotator will query for the active labels $\actlab_t$ with the input of $\overline{\pre}_t$.
Subsequently, the following loss function is adopted for the segmentation network $f_{\theta_t}$ updating,
\begin{small}
    \begin{align}
        \varL_{B1} =& \varL_{ce}(\pre_t,\actlab_t) + \varL_{ce}(\pre_t',\actlab_t) \nonumber \\
        + & \lambda_{ent}\varL_{ent}(\pre_t) + \lambda_{cst}\varL_{cst}(\pre_t,\pre_t')\,,
        \label{eq:b1}
    \end{align}
\end{small}%
where $\lambda_{ent}$ and $\lambda_{cst}$ are trade-off parameters, and $\varL_{cst}$ is the consistency regularization formulated as follows,
\begin{small}
    \begin{align}
        \varL_{cst} &=-\frac{1}{|\imgseg_t|}\sum_{(i,j)\in\imgseg_t}\sum_{c=1}^C \pre_t^{(i,j,c)}\log \pre_t'^{(i,j,c)}\,.
    \end{align}
\end{small}%

\parastart{Supervised counterparts.} In AL, there is a very important upper bound, \ie, the fully supervised approach, which trains the model with full access to ground truth in an offline manner.
Such a comparison with our online methods is meaningless.
To provide a reference, we introduce the concept of supervised counterparts, which leverage labels of all pixels, \ie, $\labseg_t$, for model adaptation after making predictions for the test image $\imgseg_t$, rather than utilizing active labels $\actlab_t$.
It is important to notice that supervised counterparts are not truly upper bounds because annotating all points during testing is impossible.
They serve as a virtual reference to demonstrate the effectiveness of our \method framework that it achieves comparable performance to full supervision with only access to extremely few annotations.

With the inclusion of label information, error accumulation is no longer a major problem, so both \Bzero and \Bone are full fine-tuning approaches.
Simultaneously, to ensure time efficiency, only one step of gradient descent is allowed for each time step $t$.

\subsection{Label Annotator}
\label{subsec:label_annotator}
Due to the extremely limited label budget, we need to determine which pixels, when annotated, will yield the maximum performance improvement for the model.
In this part, we investigate the most appropriate strategy based on the active score.
More specifically, give the pixel-level active score $\varA_t\in\Real^{H\times W}$ of the test image $\imgseg_t$, the most valuable pixel is defined as follows, 
\begin{small}
    \begin{equation}
        (i,j) = \mathop{\arg\max}_{(u,v) \notin \actlab_t} \varA_t^{(u,v)}\,,
        \label{eq:active}
    \end{equation}
\end{small}%
where the active labels $\actlab_t$ will be initialized as $\emptyset$.
Equipped with Eq.~\eqref{eq:active}, the label annotator queries labels from Oracle for the most informative pixels iteratively until the active budget $b$ is exhausted.
Next, we introduce four types of active functions to calculate the active score after obtaining the prediction $\pre_t$ of the test image.

\parastart{Random (Rand).} It selects pixels randomly from the whole image, so the active score for each pixel is sampled from a uniform distribution, expressed as $\varA_t^{(i,j)} \sim U(0,1)\,$. 

\parastart{Entropy (Ent).} In this active function, pixels with the highest predictive entropy are the most preferred, which is calculated as $\mathcal{H}_t^{(i,j)} = -\sum_{c=1}^{C}\pre_t^{(i,j,c)} \log \pre_t^{(i,j,c)}$. 
It calculates the active score by the equation $\varA_t = \mathcal{H}_t\,$.

\parastart{Region impurity and prediction uncertainty (RIPU).}
More than Ent, RIPU believes that pixels with more semantic categories in the surroundings are more valuable than ones with fewer categories.
The surrounding of each pixel $(i,j)$ in the test image is defined as $\mathcal{N}_k(i,j) = \{(u,v)||u-i|\le k, |v-j|\le k\}$.
According to prediction $\pre_t$, $\mathcal{N}_k(i,j)$ is divided in to $C$ subset $\mathcal{N}_{k,t}^c(i,j) = \{(u,v)\in \mathcal{N}_k(i,j)|\mathop{\arg\max}_{c'}\pre_t^{(u,v,c')} = c\}$.
We first compute the per-class frequency as $p(c,k,t,i,j) = {|\mathcal{N}_{k,t}^c(i,j)|}/{|\mathcal{N}_k(i,j)|}$.
Then the region impurity is formulated as $\mathcal{RP}_t^{(i,j)}=-\sum_{c=1}^C p(c,k,t,i,j) \log p(c,k,t,i,j)$
Finally, the active function is calculated by $\varA_t = \mathcal{RP}_t \odot \mathcal{H}_t$.
where $\odot$ is the element-wise matrix multiplication.

\parastart{Best vs. Second Best (BvSB).}
BvSB is another type of uncertainty estimation that the less the margin between the two highest softmax scores is, the more uncertain the pixel is. Its active score is computed as
\begin{small}
    \begin{equation}
        \varA_t^{(i,j)} = \max_c \pre_t^{(i,j,c)} - \max_{c \neq \mathop{\arg\max}_c' \pre_t^{(i,j,c')}} \pre_t^{(i,j,c)}\,.
    \end{equation}
\end{small}

\section{Experiments}
\label{sec:Experiment}

\begin{table*}[!t]
    \centering
    \scriptsize
    \caption{Experiment results (mIoU) of {\bf Cityscapes$\to$ACDC} under test protocols of both {\bf FTTA} and {\bf CTTA}. For \setting, the active {\bf budget} $b$ is 16 pixels and {\bf Fully} means the supervised counterpart. Results with flag \vidar are reported from~\cite{ViDA}, while with \datr are reported from~\cite{DAT}.}
    \label{table:ACDC}
    \VspaceBefore
    \resizebox{\textwidth}{!}
    {
        \renewcommand{\arraystretch}{0.45}
        {
            \begin{tabular}{ll|ccccc|ll|ccccc}
                \toprule[1.2pt]
                \multicolumn{7}{c|}{FTTA} & \multicolumn{2}{c}{CTTA} & \multicolumn{5}{l}{$t\xrightarrow{\hspace*{3.25cm}}$} \\
                
                \midrule
                
                TTA & Active & fog & night & rain & snow & Avg. & TTA & Active & fog & night & rain & snow & Avg. \\

                \midrule

                Source & \cblank & 67.5 & 39.7 & 61.5 & 59.0 & 56.9 & Source & \cblank & 67.5 & 39.7 & 61.5 & 59.0 & 56.9 \\
                BN & \cblank & 62.6 & 39.1 & 58.8 & 56.0 & 54.1 & BN & \cblank & 62.6 & 39.1 & 58.8 & 56.0 & 54.1 \\
                TENT & \cblank & 61.4 & 37.6 & 57.9 & 55.0 & 53.0 & TENT & \cblank & 61.4 & 37.6 & 58.2 & 55.4 & 53.2 \\
                CoTTA & \cblank & 68.5 & 40.3 & 63.1 & 60.1 & 58.0 & CoTTA &  \cblank & 68.5 & 40.3 & 63.1 & 60.0 & 58.0 \\
                DePT & \cblank  &  -&-&-&-&- & DePT\vidar & \cblank  &  71.0 & 40.8 & 58.2 & 56.8 & 56.5 \\
                VDP & \cblank  &  -&-&-&-&- & VDP\vidar & \cblank  &  70.5 & 41.1 & 62.1 & 59.5 &  58.3 \\
                DAT & \cblank  &  -&-&-&-&- & DAT\datr & \cblank  &  71.7 & 44.4 & 65.4 & 62.9 & 61.1 \\
                ViDA & \cblank  &  -&-&-&-&- & ViDA\vidar & \cblank  &  71.6 & 43.2 & 66.0 & 63.4 & 61.1 \\

                \midrule

                B0 & -  &  54.2 & 14.4 & 43.2 & 35.6 & 36.9 & B0 & - & 55.8 & 13.5 & 14.1 & 9.6 & 23.2 \\ 
                 & Rand  & 72.8 & 51.8 & 65.6 & 64.3 & 63.6 &  & Rand  & 72.8 & 52.8 & 66.9 & 66.8 & 64.8 \\ 
                 & Ent  & 70.9 & 41.6 & 64.1 & 61.1 & 59.4 &  & Ent  & 70.8 & 42.1 & 64.2 & 62.1 & 59.8 \\ 
                 & RIPU  & 73.1 & 50.7 & 66.1 & 64.4 & 63.5 &  & RIPU & 72.9 & 48.5 & 66.9 & 67.0 & 63.8 \\ 
                 & BvSB  & \bf 74.1 & \bf 53.1 & \bf 66.8 & \bf 65.3 & \bf 64.8 &  & BvSB  & \bf 74.2 & \bf 54.2 & \bf 68.2 & \bf 69.3 & \bf 66.5 \\ 
                 \rowgray & Fully  & 74.7 & 54.7 & 67.5 & 67.0 & 66.0 &  & Fully  & 74.7 & 56.1 & 68.9 & 70.5 & 67.5 \\

                 \midrule

                 B1 & -  & 64.6  & 22.4 & 54.8 & 53.3 & 48.8 & B1 & - & 64.6 & 22.3 & 34.6 & 24.9 & 36.6 \\ 
                 & Rand  & 74.1 & 53.2 & 66.3 & 65.8 & 64.9 & & Rand  &  74.1 & 54.2 & 67.1 & 67.6 & 65.7  \\ 
                 & Ent  &  71.7 & 40.6 & 64.6 & 62.0 & 59.7  &  & Ent  &  71.6 & 41.6 & 65.0 & 62.9 & 60.3 \\ 
                 & RIPU  &  73.6 & 51.6 & 66.6 & 64.9 & 64.2  &  & RIPU &  73.5 & 48.0 & 67.1 & 66.6 & 63.8  \\ 
                 & BvSB  &  \bf 75.1 & \bf 54.6 & \bf 67.3 & \bf 66.8 & \bf 65.9  &  & BvSB  & \bf 75.2 & \bf 55.0 & \bf 68.4 & \bf 69.4 & \bf 67.0  \\ 
                 \rowgray & Fully  & 75.6 & 55.8 & 68.0 & 67.8 & 66.8 &  & Fully  & 75.6 & 57.3 & 68.9 & 70.1 & 68.0 \\

                \bottomrule[1.2pt]
            \end{tabular}
        }
    }
\end{table*}
\begin{table*}[!h]
    \centering
    \vspace{-3mm}
    \caption{Experiment results (mIoU) of {\bf Cityscapes$\to$Cityscapes-C} under test protocol of {\bf FTTA}. For \setting, the active {\bf budget} $b$ is 16 pixels and {\bf Fully} means the supervised counterpart.}
    \label{table:cityscapes_full}
    \VspaceBefore
    \resizebox{\textwidth}{!}
    {
        \renewcommand{\arraystretch}{0.6}
        {
            \begin{tabular}{ll | ccccccccccccccc | c}
                \toprule[1.2pt]
                TTA & Active & \rotatebox[origin=c]{60}{motion} & \rotatebox[origin=c]{60}{snow} & \rotatebox[origin=c]{60}{fog} & \rotatebox[origin=c]{60}{shot} & \rotatebox[origin=c]{60}{defocus} & \rotatebox[origin=c]{60}{contrast} & \rotatebox[origin=c]{60}{zoom} & \rotatebox[origin=c]{60}{brightness} & \rotatebox[origin=c]{60}{frost} & \rotatebox[origin=c]{60}{elastic} & \rotatebox[origin=c]{60}{glass} & \rotatebox[origin=c]{60}{gaussian} & \rotatebox[origin=c]{60}{pixelate} & \rotatebox[origin=c]{60}{jpeg} & \rotatebox[origin=c]{60}{impulse} & Avg. \\

                \midrule
                Source & \cblank & 59.1 & 35.3 & 66.5 & 39.3 & 61.0 & 61.3 & 26.5 & 75.0 & 31.4 & 74.3 & 59.0 & 31.7 & 72.8 & 52.7 & 34.5 & 52.0 \\
                BN & \cblank & 59.1 & 36.6 & 66.1 & 40.0 & 61.1 & 61.5 & 26.9 & 74.3 & 32.8 & 73.8 & 58.7 & 32.0 & 72.5 & 52.6 & 34.6 & 52.2 \\
                TENT & \cblank & 58.1 & 35.7 & 65.1 & 39.8 & 60.3 & 60.7 & 26.1 & 73.6 & 31.8 & 73.1 & 57.7 & 31.2 & 71.7 & 52.0 & 33.4 & 51.4  \\
                CoTTA & \cblank & 59.7 & 37.4 & 67.1 & 42.1 & 61.4 & 61.3 & 26.4 & 76.3 & 31.3 & 75.7 & 59.7 & 34.0 & 74.6 & 53.9 & 35.8 & 53.1  \\

                \midrule

                B0 & - & 44.3 & 10.8 & 43.0 & 3.5 & 43.2 & 17.7 & 16.2 & 58.1 & 5.5 & 52.9 & 42.0 & 5.4 & 51.0 & 16.2 & 3.8 & 27.6  \\ 
                & Rand & 62.3 & 55.5 & 71.0 & 53.7 & 64.1 & 67.3 & 38.4 & 74.8 & 51.8 & 74.8 & 64.1 & 48.4 & 74.3 & 59.4 & 48.4 & 60.5  \\ 
                & Ent & 62.5 & 50.9 & 69.4 & 50.7 & 63.7 & 67.1 & 35.5 & 74.9 & 44.5 & 74.6 & 64.2 & 45.7 & 74.0 & 58.4 & 45.8 & 58.8 \\ 
                & RIPU & 64.2 & 56.4 & 71.1 & 54.1 & \bf 65.1 & 68.7 & 39.9 & 75.6 & 51.9 & \bf 75.7 & 66.0 & 48.0 & 75.0 & 61.1 & 48.4 & 61.4 \\ 
                & BvSB & \bf 64.4 & \bf 56.9 & \bf 72.0 & \bf 54.8 & 65.0 & \bf 68.8 & \bf 40.9 & \bf 75.9 & \bf 53.6 & 75.4 & \bf 66.3 & \bf 49.5 & \bf 75.3 & \bf 61.8 & \bf 49.5 & \bf 62.0 \\ 
                \rowgray & Fully & 64.7 & 57.9 & 72.1 & 55.5 & 66.1 & 68.9 & 42.1 & 75.8 & 55.4 & 75.7 & 66.5 & 50.7 & 75.2 & 61.9 & 51.1 & 62.6 \\ 

                \midrule

                B1 & - & 46.4 & 14.2 & 45.8 & 5.4 & 48.1 & 23.6 & 23.3 & 60.1 & 10.2 & 54.1 & 45.1 & 6.8 & 58.1 & 19.7 & 2.7 & 30.9  \\ 
                & Rand & 61.8 & 56.9 & 71.7 & 55.0 & 64.8 & 68.4 & 39.0 & 75.7 & 52.8 & 75.7 & 64.8 & 49.9 & 74.9 & 61.1 & 49.6 & 61.5  \\ 
                & Ent & 61.5 & 53.6 & 70.3 & 53.1 & 64.0 & 67.6 & 35.6 & 75.5 & 45.4 & 74.8 & 63.6 & 46.7 & 74.8 & 58.9 & 46.5 & 59.5 \\ 
                & RIPU & \bf 63.1 & 57.4 & 71.8 & 55.2 & 65.6 & 69.4 & 40.3 & 76.1 & 52.8 & 76.0 & 66.3 & 50.0 & 75.6 & 61.7 & 49.6 & 62.1 \\ 
                & BvSB & 62.7 & \bf 58.6 & \bf 72.7 & \bf 56.1 & \bf 65.9 & \bf 69.6 & \bf 41.4 & \bf 76.4 & \bf 54.3 & \bf 76.4 & \bf 66.5 & \bf 51.2 & \bf 75.8 & \bf 62.0 & \bf 50.7 & \bf 62.7 \\ 
                \rowgray & Fully & 63.5 & 59.4 & 72.6 & 56.3 & 66.6 & 69.6 & 42.4 & 76.3 & 55.8 & 76.2 & 66.8 & 51.6 & 75.7 & 62.7 & 51.7 & 63.2 \\

                \bottomrule[1.2pt]
            \end{tabular}
        }
    }
    \vspace{-0.5mm}
\end{table*}
\begin{table*}[!h]
    \centering
    \vspace{-3mm}
    \caption{Experiment results (mIoU) of {\bf Cityscapes$\to$Cityscapes-C} under test protocol of {\bf CTTA}. For \setting, the active {\bf budget} $b$ is 16 pixels and {\bf Fully} means the supervised counterpart.}
    \label{table:cityscapes_continual}
    \VspaceBefore
    \resizebox{\textwidth}{!}
    {
        \renewcommand{\arraystretch}{0.6}
        {
            \begin{tabular}{ll | ccccccccccccccc | c}
                \toprule[1.2pt]
                &Time & \multicolumn{15}{l|}{$t\xrightarrow{\hspace*{15.5cm}}$}& \\ 
                \hline
                TTA & Active & \rotatebox[origin=c]{60}{motion} & \rotatebox[origin=c]{60}{snow} & \rotatebox[origin=c]{60}{fog} & \rotatebox[origin=c]{60}{shot} & \rotatebox[origin=c]{60}{defocus} & \rotatebox[origin=c]{60}{contrast} & \rotatebox[origin=c]{60}{zoom} & \rotatebox[origin=c]{60}{brightness} & \rotatebox[origin=c]{60}{frost} & \rotatebox[origin=c]{60}{elastic} & \rotatebox[origin=c]{60}{glass} & \rotatebox[origin=c]{60}{gaussian} & \rotatebox[origin=c]{60}{pixelate} & \rotatebox[origin=c]{60}{jpeg} & \rotatebox[origin=c]{60}{impulse} & Avg. \\

                \midrule
                Source & \cblank & 59.1 & 35.3 & 66.5 & 39.3 & 61.0 & 61.3 & 26.5 & 75.0 & 31.4 & 74.3 & 59.0 & 31.7 & 72.8 & 52.7 & 34.5 & 52.0 \\
                BN & \cblank & 59.1 & 36.6 & 66.1 & 40.0 & 61.1 & 61.5 & 26.9 & 74.3 & 32.8 & 73.8 & 58.7 & 32.0 & 72.5 & 52.6 & 34.6 & 52.2 \\
                TENT & \cblank & 58.1 & 35.5 & 64.8 & 39.3 & 60.1 & 60.0 & 25.8 & 73.2 & 31.7 & 72.6 & 56.6 & 30.3 & 70.7 & 50.9 & 32.2 & 50.8  \\
                CoTTA & \cblank & 59.7 & 37.3 & 67.0 & 42.1 & 61.4 & 61.1 & 26.3 & 76.2 & 31.2 & 75.6 & 59.6 & 33.7 & 74.4 & 53.6 & 35.2 & 53.0  \\

                \midrule

                B0 & - & 43.0 & 5.0 & 10.3 & 1.7 & 18.8 & 5.5 & 2.7 & 1.9 & 1.4 & 1.9 & 1.8 & 1.4 & 1.9 & 1.9 & 1.4 & 6.7  \\ 
                & Rand & 62.3 & 55.3 & 70.1 & 56.2 & 67.8 & 69.7 & 42.7 & 76.1 & 56.5 & 77.1 & 69.6 & 54.4 & 76.2 & 65.4 & 57.9 & 63.8  \\ 
                & Ent & 62.0 & 51.3 & 69.1 & 52.1 & 64.9 & 66.1 & 36.8 & 74.0 & 46.3 & 73.2 & 64.6 & 45.8 & 72.2 & 57.5 & 48.4 & 59.0 \\ 
                & RIPU & 64.0 & 56.0 & 71.7 & 56.3 & 68.9 & 69.8 & 43.5 & 76.3 & 55.3 & 76.9 & 70.1 & 53.3 & 76.9 & 64.3 & 56.1 & 64.0 \\ 
                & BvSB & \bf 64.6 & \bf 57.5 & \bf 72.9 & \bf 57.8 & \bf 70.0 & \bf 72.4 & \bf 45.1 & \bf 78.0 & \bf 59.4 & \bf 79.1 & \bf 72.6 & \bf 56.3 & \bf 79.0 & \bf 67.7 & \bf 60.3 & \bf 66.2 \\ 
                \rowgray & Fully & 64.7 & 58.6 & 73.6 & 59.9 & 71.1 & 73.6 & 48.0 & 78.8 & 62.1 & 79.8 & 73.7 & 59.1 & 79.8 & 70.2 & 63.5 & 67.8 \\ 

                \midrule

                B1 & - & 54.8 & 9.6 & 20.5 & 4.7 & 17.9 & 16.8 & 7.5 & 14.3 & 3.6 & 5.8 & 3.2 & 3.1 & 3.1 & 3.1 & 3.1 & 11.4  \\ 
                & Rand & 61.8 & 57.0 & 71.8 & 57.5 & 68.4 & 70.7 & 43.2 & 76.6 & 57.6 & 77.5 & 70.0 & 55.7 & 76.8 & 66.1 & 58.7 & 64.6  \\ 
                & Ent & 61.7 & 52.1 & 69.8 & 52.5 & 65.2 & 66.7 & 37.5 & 73.9 & 47.5 & 73.3 & 64.4 & 46.7 & 71.8 & 56.6 & 48.6 & 59.2 \\ 
                & RIPU & \bf 63.2 & 57.5 & 72.3 & 56.9 & 69.2 & 70.0 & 42.8 & 76.4 & 55.9 & 76.8 & 70.1 & 53.9 & 77.1 & 64.8 & 57.0 & 64.3 \\ 
                & BvSB & 62.9 & \bf 59.6 & \bf 73.8 & \bf 59.6 & \bf 70.7 & \bf 73.0 & \bf 46.1 & \bf 78.4 & \bf 60.3 & \bf 79.2 & \bf 72.8 & \bf 58.2 & \bf 79.5 & \bf 68.8 & \bf 61.7 & \bf 67.0 \\ 
                \rowgray & Fully & 63.5 & 60.4 & 74.2 & 60.7 & 71.4 & 73.8 & 47.9 & 79.1 & 62.8 & 79.9 & 73.7 & 60.2 & 79.9 & 70.6 & 64.0 & 68.1 \\ 
                
                \bottomrule[1.2pt]
            \end{tabular}
        }
    }
    \vspace{-3mm}
\end{table*}

\subsection{Setup}
{\bf ACDC}~\cite{ACDC} is collected under four adverse weather, \ie, Fog, Night, Rain and Snow. {\bf Cityscapes-C} is created by applying 15 corruptions to the clean validation set of Cityscapes~\cite{cityscapes}. For more details, please refer to Appendix.

\parastart{Comparisons Methods.}
\method is compared with multiple baselines including Source, BN~\cite{BN_Stat}, Tent~\cite{tent_wang2020}, CoTTA~\cite{cotta}, DePT~\cite{DePT}, VDP~\cite{VDP}, DAT~\cite{DAT} and ViDA~\cite{ViDA}. For more details, please refer to Appendix.

\parastart{Implementation details.}
All of our experiments are conducted with PyTorch~\cite{paszke2019pytorch} framework.
For the source model, we adopt the Segformer-B5~\cite{XieWYAAL21SegFormer} pre-trained on {Cityscapes} for both tasks of transferring to {ACDC} and {Cityscapes-C}.
All the test streams are generated according to the FTTA~\cite{tent_wang2020} or CoTTA~\cite{cotta} test protocols.
Following \cite{cotta}, test images are resized to the resolution of $960\times 540$ as the inputs of the segmentation network and the predictions are evaluated under the original resolution.
For optimization, we adopt Adam~\cite{Adam} optimizer with learning rate $6.0 \times 10^{-5} / 8\text{, } \beta_1 = 0.9 \text{ and } \beta_2=0.999$. For all methods, the batch size is set to be 1. 
Concerning the hyperparameters, we adopt a unified set of values for \method across all experiments without additional claims, including $\lambda_{ent}=1.0$, $\lambda_{cst} = 1.0$ and $b=16$ pixels.

\begin{figure*}[!h]
    \centering
    \includegraphics[width=1.0\linewidth]{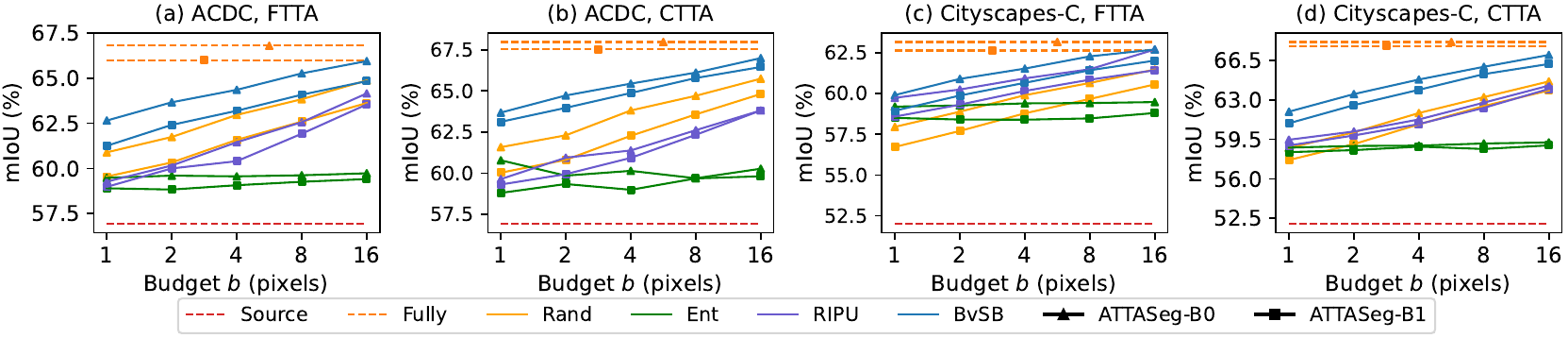}
    \vspaceBeforCaption
    \vspace{-5mm}
    \caption{{Results for ATAseg on ACDC and Cityscapes-C under test protocols of FTTA and CTTA with varying active budget $b$.}}
    \label{fig:budget}
    \vspaceAfterCaption
    \vspace{-3mm}
\end{figure*}

\subsection{Main Results}
\parastart{Generalization under domain gaps.}
Firstly, we evaluate our \method under adverse weather conditions, and the results are reported in \tab~\ref{table:ACDC}.
When focusing on superior performance, equipped with the BvSB label annotator, \Bone achieves an excellent performance under both FTTA and CTTA test protocols with only 16 pixels annotated, yielding average mIoU of $65.9\%$ and $67.0\%$ respectively, verifying the effectiveness of \method.

In more detail, it's noteworthy that BN and TENT, which demonstrate good performance in classification~\cite{BN_Stat,tent_wang2020, DePT}, exhibit the exact opposite performance in segmentation, \ie, they are inferior to the source model.
This observation holds even for DePT, which incorporates a carefully designed prompt-tuning algorithm, implying that some human empirical or heuristic intuitions in TTA may fail when transferring across tasks.
Besides, when focusing on CTTA, we can observe the supervised counterpart of \Bone outperforms the SOTA TTA methods, \ie, DAT and ViDA, by a large margin of $6.9\%$.
The above two problems increase the risk of applying TTA methods when having high security and accuracy requirements.
In contrast, \Bone surpasses all the SOTA methods, delivering $5.9\%$ improvement with extremely few annotations, making a better tradeoff between performance and cost for real-world applications.

In addition, when combined with the same label annotator, \Bone achieves consistent improvement compared to \Bzero across FTTA and CTTA, demonstrating that a superior model adapter can further enhance performance. 
Meanwhile, when adopting the same model adapter, BvSB outperforms all other types of active functions, \ie, Rand, Ent and RIPU, indicating that labeling more informative pixels will contribute to the improvement of performance (more analysis in Section~\ref{sec:ablation_study}).
To sum up, progress in either the model adapter or the label annotator can bring improvements to the \method framework.

Last, but not least, it's worth noting that both \Bzero and \Bone obtain a poor mIoU when adapting the model without any label information, which worsens as the distribution continually changes, highlighting the significant issue of error accumulation in TTA. 
Surprisingly, using the Rand label annotator, which randomly annotates 16 pixels, outperforms all existing TTA methods, demonstrating that the \method framework effectively addresses the challenges of error accumulation and determining trainable parameters by incorporating a few label information.

\parastart{Robustness under corruptions.}
We also verify the effectiveness of \method on Cityscapes-C, as reported in Table \ref{table:cityscapes_full} and \ref{table:cityscapes_continual}. 
Consistent with the previous observation,
\Bone-BvSB obtains significant performance gains compared to TTA methods, \ie, $9.6\%$ and $14.0\%$ under FTTA and CTTA respectively.
In addition, during long-term test-time adaptation, the effect of error accumulation will be even more significant, as we can see that the mIoU of \Bzero and \Bone without label annotator drops continually when performing adaptation and finally the model collapses (mIoU $\le 5\%$).
With only a few annotated pixels, our \method addresses the problem of error accumulation effectively.
Impressive performance on Cityscapes-C proves the superiority of the \method framework again.

\begin{figure*}[t]
    \centering
    \includegraphics[width=1.0\linewidth]{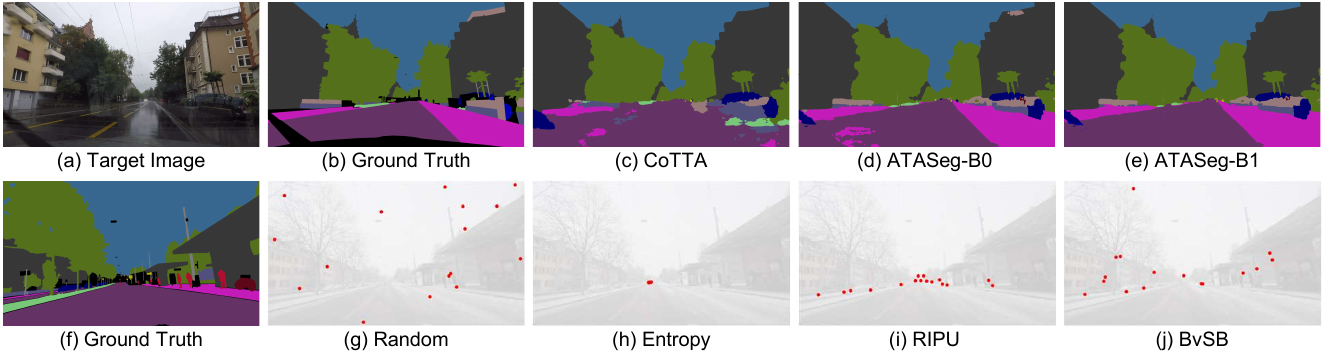}
    \vspaceBeforCaption
    \vspace{-6mm}
    \caption{{\bf Visualization}. Top row: segmentation results of different methods, and BvSB is adopted for both \Bzero and \Bone. Bottom row: actively selected pixels of different label annotators, and \Bone is employed as the model adapter for all of them.}
    \label{fig:visualization}
    \vspaceAfterCaption
\end{figure*}

\parastart{Results with varying budgets.} 
We explore the impact of varying budgets and compare the performance with the baseline method and their supervised counterparts, as illustrated in \fig~\ref{fig:budget}.
Firstly, the most obvious tendency is that with more and more pixels being annotated, both \Bzero and \Bone show consistent advancements when combined with almost all label annotators except Ent. 
This is attributed to that pixels selected by Ent always aggregate together, leading to a waste of label budgets (Section~\ref{sec:ablation_study}).
In addition, we can observe that with the same label annotator, \Bone always surpasses \Bzero across all experiments, demonstrating that a better model adapter indeed promotes the \method framework.
Meanwhile, a similar phenomenon occurs when focusing on varying label annotators with the same model adapter.
Finally, the best combination, \ie, \Bone-BvSB, consistently achieves top performance across all testing scenarios and closely approaches the performance of fully supervised counterparts after obtaining annotations for 16 points.
Remarkably, even labeling 1 pixel results in an excellent performance, validating the effectiveness of our proposed \method framework.

\subsection{Further Analysis}
\label{sec:ablation_study}
\parastart{Ablation study.} 
In \tab~\ref{table:ablation}, we further investigate the efficacy of loss functions, including unsupervised  $\varL_{ent}$ and $\varL_{cst}$, as well as supervised $\varL_{ce}$. 
All experiments employ the BvSB label annotator.
As illustrated, both \Bzero and \Bone demonstrate unsatisfactory performance without label information.
After labeling only one pixel, just using the supervised loss function $\varL_{ce}$ makes surprising improvements, \ie, $36.8\%$ and $29.7\%$ average for them respectively, validating the effectiveness of annotating the most valuable pixels. 
With all the losses participating in training, we can observe enhancements in the performance of \Bzero and \Bone across all test scenarios ($0.4\sim1.0\%$), proving the necessity of both components.  
Finally, we can further boost the performance of \method by querying a few more labels within tolerance range.

\begin{table}[t]
    \centering
    \caption{Ablation study on the unsupervised and supervised losses.}
    \label{table:ablation}
    \VspaceBefore
    \resizebox{\linewidth}{!}
    {
        \renewcommand{\arraystretch}{0.6}
        {
            \begin{tabular}{l| c c c c | cc cc | c }
                \toprule[1.2pt]
                & \multirow{2}{*}{Budget} & \multirow{2}{*}{$\varL_{ent}$} &  \multirow{2}{*}{$\varL_{cst}$} & \multirow{2}{*}{$\varL_{ce}$} & \multicolumn{2}{c}{ACDC} & \multicolumn{2}{c|}{Cityscapes-C} & \multirow{2}{*}{Avg.} \\
                & &  &  &  & FTTA & CTTA & FTTA & CTTA &  \\
                \midrule
                B0-(a) & 0 & \Checkmark & - & & 36.9 & 23.2 & 27.6 & 6.7 & 23.6 \\
                B0-(b) & 1 & & - & \Checkmark & 60.8 & 62.1 & 58.5 & 60.1 & 60.4 \\
                B0-(c) & 1 & \Checkmark & -  & \Checkmark & 61.3 & 63.1 & 58.9 & 60.9 & 61.1 \\
                B0-(d) & 16 & \Checkmark & -  & \Checkmark & 64.8 & 66.5 & 62.0 & 66.2 & 64.9 \\
                \midrule
                B1-(a) & 0 & \Checkmark & \Checkmark & & 48.8 & 36.6 & 30.9 & 11.4 & 31.9 \\
                B1-(b) & 1 & & & \Checkmark & 62.3 & 63.1 & 59.5 & 61.4 & 61.6 \\
                B1-(c) & 1 & \Checkmark & \Checkmark & \Checkmark & 62.7 & 63.7 & 59.9 & 62.0 & 62.1 \\
                B1-(d) & 16 & \Checkmark & \Checkmark & \Checkmark & 65.9 & 67.0 & 62.7 & 67.0 & 65.7 \\
                \bottomrule[1.2pt]
            \end{tabular}
        }
    }
    \vspace{-3mm}
\end{table}

\parastart{Qualitative results.} 
We present the segmentation results from \method and CoTTA in \fig~\ref{fig:visualization}. 
The predictions of \Bzero and \Bone demonstrate smoother contours and fewer spurious regions compared to those of CoTTA, indicating significant performance improvements, especially for challenging classes. 
In \fig~\ref{fig:visualization}, we showcase the selected pixels by different label annotators. 
Pixels chosen by Ent tend to cluster in a small area, providing marginal performance gains with an increasing budget. 
RIPU mitigates this issue with a specialized mechanism, achieving a noteworthy performance boost. 
Rand, while avoiding clustering problems, overlooks pixel values and fails to achieve optimal performance. In contrast, both two aspects do not trouble BvSB, making it have remarkable performance. 

\begin{figure}[t]
    \vspace{-3.5mm}
    \centering
    \includegraphics[width=1.0\linewidth]{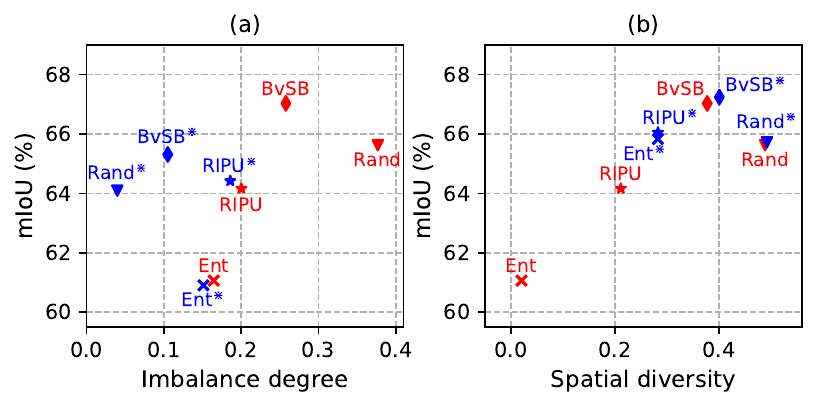}
    \vspaceBeforCaption
    \vspace{-5.5mm}
    \caption{(a) Imbalance degree and performance of label annotators. (b) Spatial diversity and performance of label annotators.}
    \label{fig:imbalance_spatial}
    \vspaceAfterCaption
    \vspace{-6mm}
\end{figure}

\parastart{Imbalance influence.}
As suggested by~\cite{RIPU}, a more balanced selection of categories will have higher mIoU.
To explore whether this criterion is valid in our framework, the frequency of each category in the selected points is counted during the adaptation process, denoted as $p_t(y)$. The imbalance degree of each label annotator is demonstrated in \fig~\ref{fig:imbalance_spatial}(a), which is calculated as $[\sum_{c=1}^C (p_N(c)- \frac{1}{C})^2]^{\frac{1}{2}}$, where $N$ is the final time step.
Among Ent, RIPU and BvSB, we observe almost the opposite conclusion that the more balanced pixels the label annotator selects, the lower mIoU it obtains.
For further comparison, we designed such a variation on all label annotators that $\varA_t^{\divideontimes(i,j)} = \varA_t^{(i,j)}\times(1 - p_{t-1}(\hat{Y}_t^{(i,j)}))$, where $\hat{Y}$ is the pseudo label.
As illustrated, while reducing the degree of imbalance, almost all methods except RIPU show performance degradation.
The above observations suggest that considering category balance in the \method may be detrimental to performance.

\parastart{Spatial diversity.}
Another very important factor affecting performance is the spatial diversity of selected pixels.
To explore its effect, we calculate the average pairwise spatial distance of annotated pixels in each image
in \fig~\ref{fig:imbalance_spatial}(b).
Firstly, when considering different values per pixel, \ie, Ent, RIPU and BvSB, the higher the spatial diversity the label annotator has, the better performance it achieves.
To further validate this observation, we enhance each label annotator with a sparsity strategy that pixels in the $129\times129$ square region centered by selected pixels are no longer selected, denoted with flag $\divideontimes$.
The simultaneous improvements in spatial diversity and performance of Ent$^\divideontimes$, RIPU$^\divideontimes$ and BvSB$^\divideontimes$ confirm this claim.
However, such a conclusion hardly holds for Rand, due to treating all pixels equally.
The lesson from the above analysis encourages increasing spatial diversity when detecting the most informative pixels.

\begin{figure}[t]
    \centering
    \vspace{-3mm}
    \includegraphics[width=1.0\linewidth]{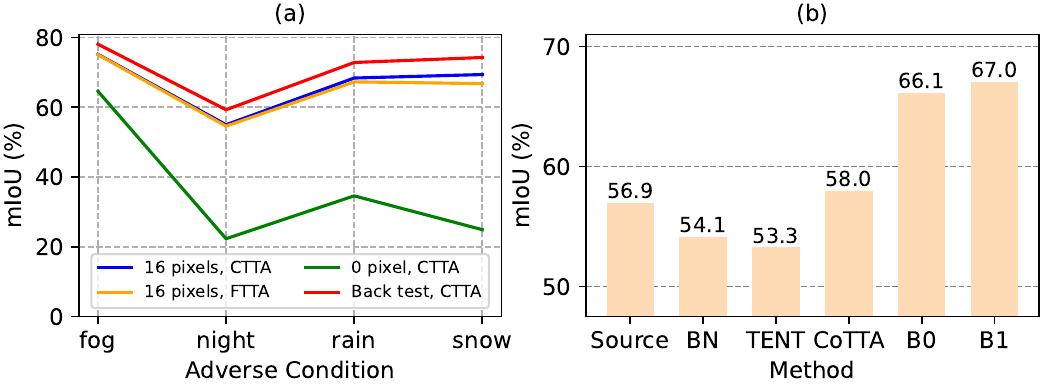}
    \vspaceBeforCaption
    \vspace{-5mm}
    \caption{(a) Catastrophic forgetting analysis. (b) Average results of different orders of ACDC under CTTA.}
    \label{fig:fogetting}
    \vspaceAfterCaption
    \vspace{-4mm}
\end{figure}

\parastart{Forgetting.} 
To explore whether ATASeg is affected by forgetting, we perform inference for all domains again in order after finishing the adaptation with \Bone-BvSB, as demonstrated in \fig~\ref{fig:fogetting}(a).
There are three points worth making: Firstly, \method is no longer bothered by the error accumulation; Secondly, continuous learning in the \method framework will contribute to future performance; Finally, catastrophic forgetting does not occur, and the performance of the historical domain is even improved.
All of these indicate the effectiveness of \method.

\parastart{Effect of the distribution changing order.} 
To exclude the effect of a fixed changing order of data distribution, we show the average results of four different orders on ACDC under the CTTA test protocol in \fig~\ref{fig:fogetting}(b).
As demonstrated, \Bzero and \Bone consistently achieve the best performance, showing the effectiveness of \method again.

\section{Conclusion}
\label{sec:conclusion}
In this paper, we comprehensively analyze the dilemma faced by TTA and identify a significant performance gap between SOTA TTA approaches and supervised counterparts. 
To overcome these challenges, the active test-time adaptation for semantic segmentation is introduced, which allows few annotations during test time, aiming to achieve a trade-off between maximizing model performance and minimizing labeling costs.
Subsequently, we propose a simple but effective \method framework, consisting of two parts, \ie, model adapter and label annotator. 
ATASeg exhibits strong compatibility, and advancements in either the model adapter or the label annotator will lead to improvements. 
Extensive experiments demonstrate that ATASeg addresses TTA dilemma and bridges the aforementioned performance gap successfully with very few annotations. 
We hope this work will open up new
ways for adapting models in real-world scenarios with requirements of high security or accuracy.  

{\small
\bibliographystyle{ieeenat_fullname}
\bibliography{reference}

\begin{thebibliography}{78}
\providecommand{\natexlab}[1]{#1}
\providecommand{\url}[1]{\texttt{#1}}
\expandafter\ifx\csname urlstyle\endcsname\relax
  \providecommand{\doi}[1]{doi: #1}\else
  \providecommand{\doi}{doi: \begingroup \urlstyle{rm}\Url}\fi

\bibitem[Ash et~al.(2020)Ash, Zhang, Krishnamurthy, Langford, and
  Agarwal]{BADGE_2020_ICLR}
Jordan~T. Ash, Chicheng Zhang, Akshay Krishnamurthy, John Langford, and Alekh
  Agarwal.
\newblock Deep batch active learning by diverse, uncertain gradient lower
  bounds.
\newblock In \emph{{ICLR}}, 2020.

\bibitem[Ban et~al.(2022)Ban, Zhang, Tong, Banerjee, and He]{BanZTBH22}
Yikun Ban, Yuheng Zhang, Hanghang Tong, Arindam Banerjee, and Jingrui He.
\newblock Improved algorithms for neural active learning.
\newblock In \emph{NeurIPS}, 2022.

\bibitem[Boudiaf et~al.(2022)Boudiaf, Mueller, Ben~Ayed, and
  Bertinetto]{niid_boudiaf2022parameter}
Malik Boudiaf, Romain Mueller, Ismail Ben~Ayed, and Luca Bertinetto.
\newblock Parameter-free online test-time adaptation.
\newblock In \emph{CVPR}, pages 8344--8353, 2022.

\bibitem[Brahma and Rai(2023)]{BrahmaR23}
Dhanajit Brahma and Piyush Rai.
\newblock A probabilistic framework for lifelong test-time adaptation.
\newblock In \emph{{CVPR}}, pages 3582--3591, 2023.

\bibitem[Cacciarelli and Kulahci(2023)]{OALsurvey}
Davide Cacciarelli and Murat Kulahci.
\newblock A survey on online active learning.
\newblock \emph{CoRR}, abs/2302.08893, 2023.

\bibitem[Cai et~al.(2021)Cai, Xu, Liew, and Foo]{CaiXLF21}
Lile Cai, Xun Xu, Jun~Hao Liew, and Chuan~Sheng Foo.
\newblock Revisiting superpixels for active learning in semantic segmentation
  with realistic annotation costs.
\newblock In \emph{CVPR}, pages 10988--10997, 2021.

\bibitem[Casanova et~al.(2020)Casanova, Pinheiro, Rostamzadeh, and
  Pal]{CasanovaPRP20}
Arantxa Casanova, Pedro~O. Pinheiro, Negar Rostamzadeh, and Christopher~J. Pal.
\newblock Reinforced active learning for image segmentation.
\newblock In \emph{ICLR}, 2020.

\bibitem[Cesa{-}Bianchi et~al.(2004)Cesa{-}Bianchi, Gentile, and
  Zaniboni]{Cesa-BianchiGZ04}
Nicol{\`{o}} Cesa{-}Bianchi, Claudio Gentile, and Luca Zaniboni.
\newblock Worst-case analysis of selective sampling for linear-threshold
  algorithms.
\newblock In \emph{NeurIPS}, pages 241--248, 2004.

\bibitem[Chen et~al.(2018)Chen, Zhu, Papandreou, Schroff, and
  Adam]{chen2018deeplabv3plus}
Liang{-}Chieh Chen, Yukun Zhu, George Papandreou, Florian Schroff, and Hartwig
  Adam.
\newblock Encoder-decoder with atrous separable convolution for semantic image
  segmentation.
\newblock In \emph{ECCV}, pages 833--851, 2018.

\bibitem[Choi et~al.(2022)Choi, Yang, Choi, and Yun]{ChoiYCY22}
Sungha Choi, Seunghan Yang, Seokeon Choi, and Sungrack Yun.
\newblock Improving test-time adaptation via shift-agnostic weight
  regularization and nearest source prototypes.
\newblock In \emph{ECCV}, pages 440--458, 2022.

\bibitem[Cordts et~al.(2016)Cordts, Omran, Ramos, Rehfeld, Enzweiler, Benenson,
  Franke, Roth, and Schiele]{cityscapes}
Marius Cordts, Mohamed Omran, Sebastian Ramos, Timo Rehfeld, Markus Enzweiler,
  Rodrigo Benenson, Uwe Franke, Stefan Roth, and Bernt Schiele.
\newblock The cityscapes dataset for semantic urban scene understanding.
\newblock In \emph{CVPR}, pages 3213--3223, 2016.

\bibitem[Dasgupta et~al.(2009)Dasgupta, Kalai, and Monteleoni]{DasguptaKM09}
Sanjoy Dasgupta, Adam~Tauman Kalai, and Claire Monteleoni.
\newblock Analysis of perceptron-based active learning.
\newblock \emph{J. Mach. Learn. Res.}, 10:\penalty0 281--299, 2009.

\bibitem[D{\"o}bler et~al.(2023)D{\"o}bler, Marsden, and Yang]{RMT}
Mario D{\"o}bler, Robert~A Marsden, and Bin Yang.
\newblock Robust mean teacher for continual and gradual test-time adaptation.
\newblock In \emph{CVPR}, pages 7704--7714, 2023.

\bibitem[Fu et~al.(2021)Fu, Cao, Wang, and Long]{Fu_2021_CVPR}
Bo Fu, Zhangjie Cao, Jianmin Wang, and Mingsheng Long.
\newblock Transferable query selection for active domain adaptation.
\newblock In \emph{CVPR}, pages 7272--7281, 2021.

\bibitem[Gan et~al.()Gan, Bai, Lou, Ma, Zhang, Shi, and Luo]{VDP}
Yulu Gan, Yan Bai, Yihang Lou, Xianzheng Ma, Renrui Zhang, Nian Shi, and Lin
  Luo.
\newblock Decorate the newcomers: Visual domain prompt for continual test time
  adaptation.
\newblock In \emph{{AAAI}}, pages 7595--7603.

\bibitem[Gao et~al.(2022)Gao, Shi, Zhu, Wang, Tang, Zhou, Li, and
  Metaxas]{DePT}
Yunhe Gao, Xingjian Shi, Yi Zhu, Hao Wang, Zhiqiang Tang, Xiong Zhou, Mu Li,
  and Dimitris~N. Metaxas.
\newblock Visual prompt tuning for test-time domain adaptation.
\newblock \emph{CoRR}, abs/2210.04831, 2022.

\bibitem[Gong et~al.(2022)Gong, Jeong, Kim, Kim, Shin, and Lee]{note}
Taesik Gong, Jongheon Jeong, Taewon Kim, Yewon Kim, Jinwoo Shin, and Sung{-}Ju
  Lee.
\newblock Robust continual test-time adaptation: Instance-aware {BN} and
  prediction-balanced memory.
\newblock In \emph{{NeurIPS}}, 2022.

\bibitem[Hanneke et~al.(2014)]{hanneke2014theory}
Steve Hanneke et~al.
\newblock Theory of disagreement-based active learning.
\newblock \emph{Found. Trends Mach. Learn.}, 7\penalty0 (2-3):\penalty0
  131--309, 2014.

\bibitem[Hao et~al.(2018)Hao, Lu, Zhao, Zhang, Hoi, and Miao]{HaoLZZHM18}
Shuji Hao, Jing Lu, Peilin Zhao, Chi Zhang, Steven C.~H. Hoi, and Chunyan Miao.
\newblock Second-order online active learning and its applications.
\newblock \emph{{IEEE} Trans. Knowl. Data Eng.}, 30\penalty0 (7):\penalty0
  1338--1351, 2018.

\bibitem[Hu et~al.(2021)Hu, Song, Gu, Luo, Chen, Chen, Zhang, and
  Zhang]{HuSGLCCZZ21}
Minhao Hu, Tao Song, Yujun Gu, Xiangde Luo, Jieneng Chen, Yinan Chen, Ya Zhang,
  and Shaoting Zhang.
\newblock Fully test-time adaptation for image segmentation.
\newblock In \emph{{MICCAI}}, pages 251--260, 2021.

\bibitem[Joshi et~al.(2009)Joshi, Porikli, and
  Papanikolopoulos]{BvSB_2009_CVPR}
Ajay~J. Joshi, Fatih Porikli, and Nikolaos Papanikolopoulos.
\newblock Multi-class active learning for image classification.
\newblock In \emph{CVPR}, 2009.

\bibitem[Kingma and Ba(2015)]{Adam}
Diederik~P. Kingma and Jimmy Ba.
\newblock Adam: {A} method for stochastic optimization.
\newblock In \emph{ICLR}, 2015.

\bibitem[Li et~al.(2023)Li, Yuan, Xie, and Yang]{grotta}
Shuang Li, Longhui Yuan, Binhui Xie, and Tao Yang.
\newblock Generalized robust test-time adaptation in continuous dynamic
  scenarios.
\newblock \emph{CoRR}, abs/2310.04714, 2023.

\bibitem[Liang et~al.(2023)Liang, He, and Tan]{liang2023ttasurvey}
Jian Liang, Ran He, and Tieniu Tan.
\newblock A comprehensive survey on test-time adaptation under distribution
  shifts.
\newblock \emph{CoRR}, abs/2303.15361, 2023.

\bibitem[Lim et~al.(2023)Lim, Kim, Choo, and Choi]{TTN}
Hyesu Lim, Byeonggeun Kim, Jaegul Choo, and Sungha Choi.
\newblock {TTN:} {A} domain-shift aware batch normalization in test-time
  adaptation.
\newblock In \emph{ICLR}, 2023.

\bibitem[Liu et~al.(2023)Liu, Yang, Jia, Lu, Guo, Xue, and Zhang]{ViDA}
Jiaming Liu, Senqiao Yang, Peidong Jia, Ming Lu, Yandong Guo, Wei Xue, and
  Shanghang Zhang.
\newblock Vida: Homeostatic visual domain adapter for continual test time
  adaptation.
\newblock \emph{CoRR}, abs/2306.04344, 2023.

\bibitem[Loy et~al.(2012)Loy, Hospedales, Xiang, and Gong]{LoyStream12}
Chen~Change Loy, Timothy~M. Hospedales, Tao Xiang, and Shaogang Gong.
\newblock Stream-based joint exploration-exploitation active learning.
\newblock In \emph{CVPR}, pages 1560--1567, 2012.

\bibitem[Lu et~al.(2016)Lu, Zhao, and Hoi]{LuZH16}
Jing Lu, Peilin Zhao, and Steven C.~H. Hoi.
\newblock Online passive-aggressive active learning.
\newblock \emph{Mach. Learn.}, 103\penalty0 (2):\penalty0 141--183, 2016.

\bibitem[Nado et~al.(2020)Nado, Padhy, Sculley, D'Amour, Lakshminarayanan, and
  Snoek]{BN_Stat}
Zachary Nado, Shreyas Padhy, D. Sculley, Alexander D'Amour, Balaji
  Lakshminarayanan, and Jasper Snoek.
\newblock Evaluating prediction-time batch normalization for robustness under
  covariate shift.
\newblock \emph{CoRR}, abs/2006.10963, 2020.

\bibitem[Ni et~al.(2023)Ni, Yang, Liu, Li, Jiao, Xu, Chen, Liu, and Zhang]{DAT}
Jiayi Ni, Senqiao Yang, Jiaming Liu, Xiaoqi Li, Wenyu Jiao, Ran Xu, Zehui Chen,
  Yi Liu, and Shanghang Zhang.
\newblock Distribution-aware continual test time adaptation for semantic
  segmentation.
\newblock \emph{CoRR}, abs/2309.13604, 2023.

\bibitem[Niloy et~al.(2023)Niloy, Ahmed, Raychaudhuri, Oymak, and
  Roy{-}Chowdhury]{NiloyTTA2023}
Fahim~Faisal Niloy, Sk~Miraj Ahmed, Dripta~S. Raychaudhuri, Samet Oymak, and
  Amit~K. Roy{-}Chowdhury.
\newblock Effective restoration of source knowledge in continual test time
  adaptation.
\newblock \emph{CoRR}, abs/2311.04991, 2023.

\bibitem[Ning et~al.(2021)Ning, Lu, Wei, Bian, Yuan, Yu, Ma, and
  Zheng]{NingL0BYY0021}
Munan Ning, Donghuan Lu, Dong Wei, Cheng Bian, Chenglang Yuan, Shuang Yu, Kai
  Ma, and Yefeng Zheng.
\newblock Multi-anchor active domain adaptation for semantic segmentation.
\newblock In \emph{{ICCV}}, pages 9092--9102, 2021.

\bibitem[Niu et~al.(2022)Niu, Wu, Zhang, Chen, Zheng, Zhao, and Tan]{EATA}
Shuaicheng Niu, Jiaxiang Wu, Yifan Zhang, Yaofo Chen, Shijian Zheng, Peilin
  Zhao, and Mingkui Tan.
\newblock Efficient test-time model adaptation without forgetting.
\newblock In \emph{ICML}, pages 16888--16905, 2022.

\bibitem[Niu et~al.(2023)Niu, Wu, Zhang, Wen, Chen, Zhao, and Tan]{Niu00WCZT23}
Shuaicheng Niu, Jiaxiang Wu, Yifan Zhang, Zhiquan Wen, Yaofo Chen, Peilin Zhao,
  and Mingkui Tan.
\newblock Towards stable test-time adaptation in dynamic wild world.
\newblock In \emph{ICLR}, 2023.

\bibitem[Park et~al.(2023)Park, Kim, Kwon, Yoon, and Sohn]{ParkTTA2023}
Junyoung Park, Jin Kim, Hyeongjun Kwon, Ilhoon Yoon, and Kwanghoon Sohn.
\newblock Layer-wise auto-weighting for non-stationary test-time adaptation.
\newblock \emph{CoRR}, abs/2311.05858, 2023.

\bibitem[Paszke et~al.(2019)Paszke, Gross, Massa, Lerer, Bradbury, Chanan,
  Killeen, Lin, Gimelshein, Antiga, et~al.]{paszke2019pytorch}
Adam Paszke, Sam Gross, Francisco Massa, Adam Lerer, James Bradbury, Gregory
  Chanan, Trevor Killeen, Zeming Lin, Natalia Gimelshein, Luca Antiga, et~al.
\newblock Pytorch: An imperative style, high-performance deep learning library.
\newblock In \emph{NeurIPS}, pages 8024--8035, 2019.

\bibitem[Prabhu et~al.(2021)Prabhu, Chandrasekaran, Saenko, and
  Hoffman]{CULE_2021_ICCV}
Viraj Prabhu, Arjun Chandrasekaran, Kate Saenko, and Judy Hoffman.
\newblock Active domain adaptation via clustering uncertainty-weighted
  embeddings.
\newblock In \emph{ICCV}, pages 8505--8514, 2021.

\bibitem[Quionero-Candela et~al.(2009)Quionero-Candela, Sugiyama, Schwaighofer,
  and Lawrence]{dataset_shift_in_ML09}
Joaquin Quionero-Candela, Masashi Sugiyama, Anton Schwaighofer, and Neil~D.
  Lawrence.
\newblock \emph{Dataset Shift in Machine Learning}.
\newblock The MIT Press, 2009.

\bibitem[Rai et~al.(2010)Rai, Saha, Daum{\'e}~III, and
  Venkatasubramanian]{rai2010domain}
Piyush Rai, Avishek Saha, Hal Daum{\'e}~III, and Suresh Venkatasubramanian.
\newblock Domain adaptation meets active learning.
\newblock In \emph{ALNLP Workshop}, pages 27--32, 2010.

\bibitem[Ren et~al.(2022)Ren, Xiao, Chang, Huang, Li, Gupta, Chen, and
  Wang]{RenXCHLGCW22}
Pengzhen Ren, Yun Xiao, Xiaojun Chang, Po{-}Yao Huang, Zhihui Li, Brij~B.
  Gupta, Xiaojiang Chen, and Xin Wang.
\newblock A survey of deep active learning.
\newblock \emph{{ACM} Comput. Surv.}, 54\penalty0 (9):\penalty0 180:1--180:40,
  2022.

\bibitem[Ronneberger et~al.(2015)Ronneberger, Fischer, and
  Brox]{ronneberger2015u-net}
Olaf Ronneberger, Philipp Fischer, and Thomas Brox.
\newblock U-net: Convolutional networks for biomedical image segmentation.
\newblock In \emph{MICCAI}, pages 234--241, 2015.

\bibitem[Sakaridis et~al.(2021)Sakaridis, Dai, and Van~Gool]{ACDC}
Christos Sakaridis, Dengxin Dai, and Luc Van~Gool.
\newblock Acdc: The adverse conditions dataset with correspondences for
  semantic driving scene understanding.
\newblock In \emph{ICCV}, pages 10765--10775, 2021.

\bibitem[Saran et~al.(2023)Saran, Yousefi, Krishnamurthy, Langford, and
  Ash]{SaranYK0A23}
Akanksha Saran, Safoora Yousefi, Akshay Krishnamurthy, John Langford, and
  Jordan~T. Ash.
\newblock Streaming active learning with deep neural networks.
\newblock In \emph{{ICML}}, pages 30005--30021, 2023.

\bibitem[Sener and Savarese(2018)]{CoreSet_2019_ICLR}
Ozan Sener and Silvio Savarese.
\newblock Active learning for convolutional neural networks: {A} core-set
  approach.
\newblock In \emph{ICLR}, 2018.

\bibitem[Settles(2009)]{settles2009active}
Burr Settles.
\newblock Active learning literature survey.
\newblock 2009.

\bibitem[Shah and Manwani(2020)]{ShahM20}
Kulin Shah and Naresh Manwani.
\newblock Online active learning of reject option classifiers.
\newblock In \emph{{AAAI}}, pages 5652--5659, 2020.

\bibitem[Shan et~al.(2019)Shan, Zhang, Liu, and Liu]{ShanALEnsemble}
Jicheng Shan, Hang Zhang, Weike Liu, and Qingbao Liu.
\newblock Online active learning ensemble framework for drifted data streams.
\newblock \emph{Trans. Neural Networks Learn. Syst.}, 30\penalty0 (2):\penalty0
  486--498, 2019.

\bibitem[Sheikh et~al.(2020)Sheikh, Milioto, Lottes, Stachniss, Bennewitz, and
  Schultz]{SheikhMLSBS20}
Rasha Sheikh, Andres Milioto, Philipp Lottes, Cyrill Stachniss, Maren
  Bennewitz, and Thomas Schultz.
\newblock Gradient and log-based active learning for semantic segmentation of
  crop and weed for agricultural robots.
\newblock In \emph{ICRA}, pages 1350--1356, 2020.

\bibitem[Shen et~al.(2018)Shen, Yun, Lipton, Kronrod, and Anandkumar]{ENT}
Yanyao Shen, Hyokun Yun, Zachary~C. Lipton, Yakov Kronrod, and Animashree
  Anandkumar.
\newblock Deep active learning for named entity recognition.
\newblock In \emph{{ICLR}}, 2018.

\bibitem[Shin et~al.(2021)Shin, Kim, Cho, Woo, Park, and Kweon]{Shin0CWPK21}
Inkyu Shin, Dong{-}Jin Kim, Jae{-}Won Cho, Sanghyun Woo, KwanYong Park, and
  In~So Kweon.
\newblock Labor: Labeling only if required for domain adaptive semantic
  segmentation.
\newblock In \emph{{ICCV}}, pages 8568--8578, 2021.

\bibitem[Siddiqui et~al.(2020)Siddiqui, Valentin, and
  Nie{\ss}ner]{siddiqui_2020_CVPR}
Yawar Siddiqui, Julien Valentin, and Matthias Nie{\ss}ner.
\newblock Viewal: Active learning with viewpoint entropy for semantic
  segmentation.
\newblock In \emph{{CVPR}}, pages 9430--9440, 2020.

\bibitem[Sinha et~al.(2019)Sinha, Ebrahimi, and Darrell]{sinha2019variational}
Samarth Sinha, Sayna Ebrahimi, and Trevor Darrell.
\newblock Variational adversarial active learning.
\newblock In \emph{ICCV}, pages 5972--5981, 2019.

\bibitem[Song et~al.(2023)Song, Lee, Kweon, and Choi]{song2023ecotta}
Junha Song, Jungsoo Lee, In~So Kweon, and Sungha Choi.
\newblock Ecotta: Memory-efficient continual test-time adaptation via
  self-distilled regularization.
\newblock In \emph{CVPR}, pages 11920--11929, 2023.

\bibitem[Strudel et~al.(2021)Strudel, Pinel, Laptev, and Schmid]{StrudelPLS21}
Robin Strudel, Ricardo~Garcia Pinel, Ivan Laptev, and Cordelia Schmid.
\newblock Segmenter: Transformer for semantic segmentation.
\newblock In \emph{{ICCV}}, pages 7242--7252, 2021.

\bibitem[Su et~al.(2020)Su, Tsai, Sohn, Liu, Maji, and Chandraker]{AADA_WACV}
Jong{-}Chyi Su, Yi{-}Hsuan Tsai, Kihyuk Sohn, Buyu Liu, Subhransu Maji, and
  Manmohan Chandraker.
\newblock Active adversarial domain adaptation.
\newblock In \emph{{WACV}}, pages 728--737, 2020.

\bibitem[Taghanaki et~al.(2021)Taghanaki, Abhishek, Cohen, Cohen{-}Adad, and
  Hamarneh]{TaghanakiACCH21}
Saeid~Asgari Taghanaki, Kumar Abhishek, Joseph~Paul Cohen, Julien Cohen{-}Adad,
  and Ghassan Hamarneh.
\newblock Deep semantic segmentation of natural and medical images: a review.
\newblock \emph{Artif. Intell. Rev.}, 54\penalty0 (1):\penalty0 137--178, 2021.

\bibitem[Teichmann et~al.(2018)Teichmann, Weber, Z{\"{o}}llner, Cipolla, and
  Urtasun]{TeichmannWZCU18}
Marvin Teichmann, Michael Weber, J.~Marius Z{\"{o}}llner, Roberto Cipolla, and
  Raquel Urtasun.
\newblock Multinet: Real-time joint semantic reasoning for autonomous driving.
\newblock In \emph{IV}, pages 1013--1020, 2018.

\bibitem[Valada et~al.(2017)Valada, Vertens, Dhall, and Burgard]{ValadaVDB17}
Abhinav Valada, Johan Vertens, Ankit Dhall, and Wolfram Burgard.
\newblock Adapnet: Adaptive semantic segmentation in adverse environmental
  conditions.
\newblock In \emph{ICRA}, pages 4644--4651, 2017.

\bibitem[Wang and Shang(2014)]{entropy_2014_IJCNN}
Dan Wang and Yi Shang.
\newblock A new active labeling method for deep learning.
\newblock In \emph{{IJCNN}}, pages 112--119, 2014.

\bibitem[Wang et~al.(2021{\natexlab{a}})Wang, Shelhamer, Liu, Olshausen, and
  Darrell]{tent_wang2020}
Dequan Wang, Evan Shelhamer, Shaoteng Liu, Bruno~A. Olshausen, and Trevor
  Darrell.
\newblock Tent: Fully test-time adaptation by entropy minimization.
\newblock In \emph{{ICLR}}, 2021{\natexlab{a}}.

\bibitem[Wang et~al.(2022)Wang, Fink, Gool, and Dai]{cotta}
Qin Wang, Olga Fink, Luc~Van Gool, and Dengxin Dai.
\newblock Continual test-time domain adaptation.
\newblock In \emph{{CVPR}}, pages 7191--7201, 2022.

\bibitem[Wang et~al.(2023{\natexlab{a}})Wang, Zhang, Yan, Zhang, and
  Li]{WangZYZL23}
Shuai Wang, Daoan Zhang, Zipei Yan, Jianguo Zhang, and Rui Li.
\newblock Feature alignment and uniformity for test time adaptation.
\newblock In \emph{CVPR}, pages 20050--20060, 2023{\natexlab{a}}.

\bibitem[Wang et~al.(2021{\natexlab{b}})Wang, Awasthi, Dann, Sekhari, and
  Gentile]{WangADSG21}
Zhilei Wang, Pranjal Awasthi, Christoph Dann, Ayush Sekhari, and Claudio
  Gentile.
\newblock Neural active learning with performance guarantees.
\newblock In \emph{NeurIPS}, pages 7510--7521, 2021{\natexlab{b}}.

\bibitem[Wang et~al.(2023{\natexlab{b}})Wang, Luo, Zheng, Chen, Wang, and
  Huang]{Wang23ttasurvey}
Zixin Wang, Yadan Luo, Liang Zheng, Zhuoxiao Chen, Sen Wang, and Zi Huang.
\newblock In search of lost online test-time adaptation: {A} survey.
\newblock \emph{CoRR}, abs/2310.20199, 2023{\natexlab{b}}.

\bibitem[Wu et~al.(2021)Wu, Liu, Huang, Lee, Su, Huang, and Hsu]{Wu_2021_ICCV}
Tsung-Han Wu, Yueh-Cheng Liu, Yu-Kai Huang, Hsin-Ying Lee, Hung-Ting Su,
  Ping-Chia Huang, and Winston~H. Hsu.
\newblock Redal: Region-based and diversity-aware active learning for point
  cloud semantic segmentation.
\newblock In \emph{ICCV}, pages 15510--15519, 2021.

\bibitem[Xie et~al.(2022{\natexlab{a}})Xie, Yuan, Li, Liu, and Cheng]{RIPU}
Binhui Xie, Longhui Yuan, Shuang Li, Chi~Harold Liu, and Xinjing Cheng.
\newblock Towards fewer annotations: Active learning via region impurity and
  prediction uncertainty for domain adaptive semantic segmentation.
\newblock In \emph{{CVPR}}, pages 8058--8068, 2022{\natexlab{a}}.

\bibitem[Xie et~al.(2022{\natexlab{b}})Xie, Yuan, Li, Liu, Cheng, and
  Wang]{xie2022active}
Binhui Xie, Longhui Yuan, Shuang Li, Chi~Harold Liu, Xinjing Cheng, and Guoren
  Wang.
\newblock Active learning for domain adaptation: An energy-based approach.
\newblock In \emph{AAAI}, pages 8708--8716, 2022{\natexlab{b}}.

\bibitem[Xie et~al.(2023)Xie, Li, Guo, Liu, and Cheng]{xie2023annotator}
Binhui Xie, Shuang Li, Qingju Guo, Harold~Chi Liu, and Xinjing Cheng.
\newblock Annotator: An generic active learning baseline for lidar semantic
  segmentation.
\newblock In \emph{NeurIPS}, 2023.

\bibitem[Xie et~al.(2021)Xie, Wang, Yu, Anandkumar, {\'{A}}lvarez, and
  Luo]{XieWYAAL21SegFormer}
Enze Xie, Wenhai Wang, Zhiding Yu, Anima Anandkumar, Jos{\'{e}}~M.
  {\'{A}}lvarez, and Ping Luo.
\newblock Segformer: Simple and efficient design for semantic segmentation with
  transformers.
\newblock In \emph{NeurIPS}, pages 12077--12090, 2021.

\bibitem[Yang et~al.(2018)Yang, Yu, Zhang, Li, and Yang]{YangYZLY18}
Maoke Yang, Kun Yu, Chi Zhang, Zhiwei Li, and Kuiyuan Yang.
\newblock Denseaspp for semantic segmentation in street scenes.
\newblock In \emph{CVPR}, pages 3684--3692, 2018.

\bibitem[Yi et~al.(2023)Yi, Yang, Wang, Li, Tan, and Kot]{YiYWLTK23}
Chenyu Yi, Siyuan Yang, Yufei Wang, Haoliang Li, Yap{-}Peng Tan, and Alex~C.
  Kot.
\newblock Temporal coherent test time optimization for robust video
  classification.
\newblock In \emph{{ICLR}}, 2023.

\bibitem[Yu et~al.(2023{\natexlab{a}})Yu, Shi, and Yu]{activetest}
Dayou Yu, Weishi Shi, and Qi Yu.
\newblock Actively testing your model while it learns: Realizing
  label-efficient learning in practice.
\newblock In \emph{NeurIPS}, 2023{\natexlab{a}}.

\bibitem[Yu et~al.(2023{\natexlab{b}})Yu, Li, Du, Li, Zhu, and Yang]{YuLDLZY23}
Zhiqi Yu, Jingjing Li, Zhekai Du, Fengling Li, Lei Zhu, and Yang Yang.
\newblock Noise-robust continual test-time domain adaptation.
\newblock In \emph{ACM MM}, pages 2654--2662, 2023{\natexlab{b}}.

\bibitem[Yuan et~al.(2023)Yuan, Xie, and Li]{rotta}
Longhui Yuan, Binhui Xie, and Shuang Li.
\newblock Robust test-time adaptation in dynamic scenarios.
\newblock In \emph{CVPR}, pages 15922--15932, 2023.

\bibitem[Zhang et~al.(2023)Zhang, Qi, Shi, and Gao]{zhang2023domainadaptor}
Jian Zhang, Lei Qi, Yinghuan Shi, and Yang Gao.
\newblock Domainadaptor: A novel approach to test-time adaptation.
\newblock In \emph{ICCV}, pages 18971--18981, 2023.

\bibitem[Zhang et~al.(2022)Zhang, Levine, and Finn]{zhang2022memo}
Marvin~Mengxin Zhang, Sergey Levine, and Chelsea Finn.
\newblock {MEMO}: Test time robustness via adaptation and augmentation.
\newblock In \emph{NeurIPS}, 2022.

\bibitem[Zhdanov(2019)]{zhdanov2019diverse}
Fedor Zhdanov.
\newblock Diverse mini-batch active learning.
\newblock \emph{CoRR}, abs/1901.05954, 2019.

\bibitem[Zhou et~al.(2023)Zhou, Guo, Jia, Zhang, and Li]{ODS}
Zhi Zhou, Lan{-}Zhe Guo, Lin{-}Han Jia, Dingchu Zhang, and Yu{-}Feng Li.
\newblock {ODS:} test-time adaptation in the presence of open-world data shift.
\newblock In \emph{ICML}, pages 42574--42588, 2023.

\end{thebibliography}
}

\vspace{5cm}
\section{Appendix}
\label{sec:appendix}
\begin{figure*}[t]
    \centering
    \includegraphics[width=1.0\textwidth]{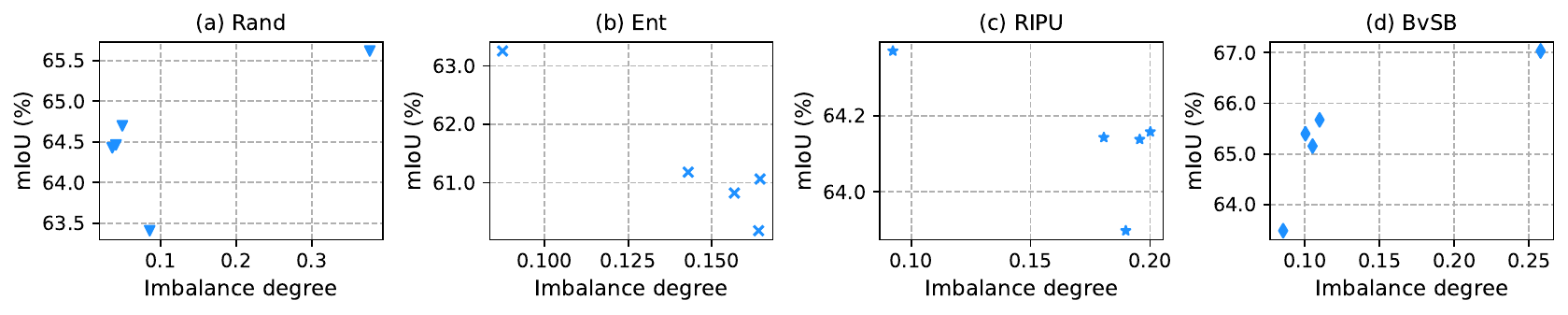}
    \vspaceBeforCaption
    \vspace{-4mm}
    \caption{Imbalance degree and performance of different enhanced label annotators on ACDC under CTTA test protocol.}
    \vspaceAfterCaption
    \label{fig:appendxi_imb}
\end{figure*}
\begin{figure*}[!h]
    \centering
    \includegraphics[width=1.0\textwidth]{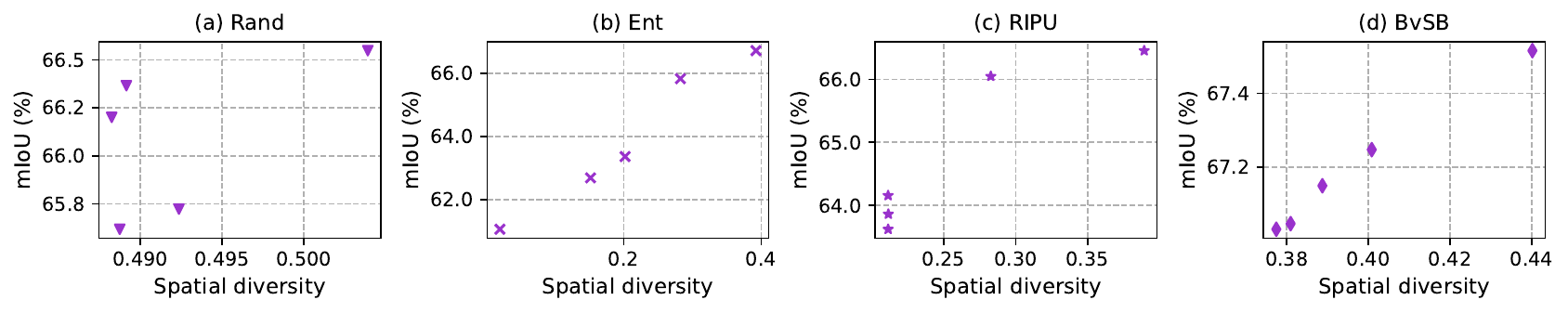}
    \vspace{-4mm}
    \vspaceBeforCaption
    \caption{Spatial Diversity and performance of different enhanced label annotators on ACDC under CTTA test protocol.}
    \label{fig:appendxi_spa}
    \vspaceAfterCaption
\end{figure*}

\subsection{Additional Experiment Details and Results}

\subsubsection{Datasets}
\parastart{ACDC}~\cite{ACDC} dataset shares $19$ classes with Cityscapes, encompassing $4006$ images with a resolution of $1920\times1080$.
These images are collected under four distinct, unseen visual conditions: Fog, Night, Rain and Snow. 
All test streams for experiments on ACDC originate with the whole training set of the mentioned domains.

\parastart{Cityscapes}~\cite{cityscapes} is a dataset of real urban scenes, comprising $2,975$ training images, $500$ validation images, and $1,525$ test images, all finely annotated and with a resolution set at $2048\times1024$. 
Each pixel of the image is categorized into one of $19$ categories. 
{\bf Cityscapes-C} dataset is created by applying 15 different types of corruption to the clean {\it validation} images of Cityscapes to assess the robustness of networks under corruption.
In our experiments, all test streams of Cityscapes-C are generated on the {\it highest severity 5} of all corruptions.
{\bf Cityscapes-foggy} dataset is a synthetic foggy dataset that simulates fog on real scenes. 
In this appendix, we report additional results on the {\it highest foggy level} under the FTTA test protocol.

\subsubsection{Compared methods}
\parastart{BN}~\cite{BN_Stat} normalizes the feature map of the current batch of samples based on their statistics, without the necessity to update any network parameters. 

\parastart{Tent}~\cite{tent_wang2020} proposes fully test-time adaptation for the first time. 
It incorporates test-time batch normalization and utilizes entropy minimization for the training of the affine parameters associated with Batch Normalization (BN) layers. 
We reimplement it following the released code \url{https://github.com/DequanWang/tent}.

\parastart{CoTTA}~\cite{cotta} introduces a setup for continually adapting a pre-trained source model to the current test data. 
To address error accumulation and catastrophic forgetting, it leverages augmentation-averaged pseudo-labels and stochastic restoration. 
We reimplement it following the released code \url{https://github.com/qinenergy/cotta}.

\parastart{DePT}~\cite{DePT} collects learnable visual prompts which are trained using labeled source data and vision Transformer. 
Subsequently, these prompts and classification heads are fine-tuned during testing.

\parastart{VDP}~\cite{VDP} learns domain-specific prompts and domain-agnostic prompts, presenting a homeostasis-based prompt adaptation strategy to acquire domain-shared knowledge. 
In the adaptation phase, test data is adapted to the source model with the assistance of the prompts, while maintaining the parameters of the source model fixed.

\parastart{DAT}~\cite{DAT} explores the task of semantic segmentation under CTTA, and proposes to select small groups of trainable domain-specific and task-relevant parameters for updating via the
parameter accumulation update strategy.

\parastart{ViDA}~\cite{ViDA} designs a homeostatic visual domain adapter with low-rank and high-rank embedding space and utilizes a homeostatic knowledge allotment strategy to combine different knowledge from each adapter.

\subsubsection{Imbalance Influence}
As mentioned in the main paper, considering category balance in the \method may be detrimental to performance. 
To investigate the impact of it, we further design variants for all label annotators that $\varA_t^{\divideontimes(i,j)} = (1-\omega)\varA_t^{(i,j)}+ \omega(1 - p_{t-1}(\hat{Y}_t^{(i,j)}))$, where $\omega$ controls the imbalance degree. 
We vary $\omega\in\{0.0,0.25,0.5,0.75,1.0\}$ and conduct experiments on ACDC under the CTTA test protocol.
The results of different label annotators with \Bone are demonstrated in \fig~\ref{fig:appendxi_imb}.
Firstly, when focusing on Rand and BvSB, we can observe that as the selected pixels become more class-balanced, the performance exhibits different degrees of decline, \ie, $1\sim4\%$.
Secondly, as for RIPU, the fluctuation contributing to category balance is small, which is less than $0.5\%$.
Finally, when the imbalance degree becomes the smallest in \fig~\ref{fig:appendxi_imb}(b), Ent obtains a significant performance gain (about $2.0\%$).
However, such an improvement does not result from integrating the category balance into the informative score.
This is because the improvement is obtained at $\omega=0$, where the score function becomes $\varA_t^{\divideontimes(i,j)} = 1 - p_{t-1}(\hat{Y}_t^{(i,j)})$, which exactly means another type of label annotator.
Based on the experimental observations so far, the gains from combining category balance are modest and even harmful.
We still need to be careful when developing algorithms considering category balance in the framework of \method.

\begin{table}
    \centering
    \scriptsize
    \caption{Different consistency loss $\varL_{cst}$ for \Bone-BvSB.}
    \label{tab:consistency}
    \vspaceAfterCaption
    \resizebox{\linewidth}{!}
    {
        \begin{tabular}{l | c c c c | c}
            \toprule[0.7pt]
            \multirow{2}{*}{$\varL_{cst}$} & \multicolumn{2}{c}{ACDC} & \multicolumn{2}{c|}{Cityscapes-C} & \multirow{2}{*}{Avg.} \\
            & FTTA & CTTA & FTTA & CTTA & \\
            \midrule
            $\varL_{l_1}$ & \bf 65.9 & 67.1 & 62.9 & 66.8 & 65.7 \\
            $\varL_{mse}$ & 65.8 & \bf 67.2 & \bf 63.2 & 66.8 & \bf 65.8 \\
            $\varL_{sce}$ & \bf 65.9 & 67.0 & 62.7 & \bf 67.0 & 65.7 \\
            \bottomrule[0.7pt]
        \end{tabular}
    }
    \vspace{-4mm}
\end{table}

\subsubsection{Spatial Diversity}
Due to the natural neighborhood similarity of the images, the aggregation of selected pixels becomes the main challenge of the score-based active strategies.
In the main paper, we develop each label annotator by a sparsity strategy that pixels in the $k\times k$ square region centered by selected pixels are no longer selected. 
The results of different label annotators with varying $k\in\{17,33,65,129,257\}$ are demonstrated in \fig~\ref{fig:appendxi_spa}, where all experiments employ \Bone.
An obvious trend is that with the increase of spatial sparsity, the performances of Ent, RIPU and BvSB are all improved, further confirming the impact of aggregation.
The deeper reason is that the sparsity strategy provides a type of implicit diversity information.
After being enhanced, Ent, RIPU and BvSB become hybrid selection strategies.
This inspires the subsequent algorithm design, \ie, integrating the uncertainty and the diversity information in a time-efficient manner.

\begin{table*}[t]
    \centering
    \caption{Per-class IoU and mIoU on the task of Cityscapes$\to$Cityscapes-foggy under test protocol of FTTA.}
    \label{tab:cityfoggy}
    \vspaceAfterCaption
    \resizebox{\textwidth}{!}
    {
        \renewcommand{\arraystretch}{1.0}
        {
            \begin{tabular}{ll | ccccccccccccccccccc | c}
                \toprule[1.2pt]
                TTA & Active & road & side. & buil. & wall & fence & pole & light & sign & veg. & terr. & sky & pers. & rider & car & truck & bus & train & mbike & bike & mIoU \\
                \midrule 
                Source & - & 97.4 & 81.2 & 83.0 & 52.9 & 53.6 & 50.3 & 51.2 & 67.6 & 80.9 & 62.2 & 68.1 & 72.4 & 56.7 & 91.5 & 77.0 & 78.4 & 70.1 & 54.5 & 69.1 & 69.4 \\
                BN & - & 97.2 & 80.8 & 82.3 & 50.8 & 53.1 & 49.9 & 51.3 & 66.5 & 80.3 & 61.5 & 68.9 & 72.1 & 56.7 & 91.1 & 75.5 & 78.0 & 69.1 & 53.6 & 68.7 & 68.8 \\
                TENT & - & 97.2 & 80.5 & 82.1 & 50.6 & 52.3 & 48.7 & 50.1 & 65.7 & 80.1 & 61.0 & 69.2 & 71.6 & 55.6 & 90.5 & 69.8 & 76.5 & 68.8 & 51.5 & 68.2 & 67.9 \\
                CoTTA & - & 97.4 & 81.3 & 83.2 & 54.5 & 55.0 & 51.5 & 52.8 & 69.0 & 81.0 & 62.3 & 68.7 & 73.5 & \bf 59.0 & 91.7 & 77.8 & 78.5 & 71.6 & 56.8 & 70.5 & 70.3 \\
                \midrule 
                B1 & BvSB & \bf 97.8 & \bf 83.3 & \bf 88.2 & \bf 57.1 & \bf 56.6 & \bf 54.4 & \bf 57.2 & \bf 70.6 & \bf 87.2 & \bf 63.1 & \bf 82.9 & \bf 75.0 & 58.6 & \bf 92.6 & \bf 78.8 & \bf 82.8 & \bf 76.8 & \bf 57.2 & \bf 72.0 & \bf 73.3 \\
                \bottomrule[1.2pt]
            \end{tabular}
        }
    }
\end{table*}

\begin{table}
    \centering
    \scriptsize
    \caption{Parameter sensitivity of \Bone-BvSB.}
    \label{tab:sensitivity}
    \vspaceAfterCaption
    \resizebox{\linewidth}{!}
    {
        \begin{tabular}{c | c c c c c}
            \toprule[0.7pt]
            \diagbox{$\varL_{ent}$}{$\varL_{cst}$} & 0.1 & 0.5 & 1.0 & 1.5 & 2.0 \\
            \hline
            0.1 &  67.0 & 67.3 & 67.1 & 67.0 & 67.1 \\
            0.5 & 67.2 & 66.7 & 66.9 & 67.3 & 66.9 \\
            1.0 & 67.0 & 67.3 & 67.0 & 66.9 & 66.9 \\
            1.5 & 67.6 & 67.1 & 67.2 & 67.2 & 67.1 \\
            2.0 & 67.2 & 66.8 & 66.8 & 66.9 & 67.0 \\
            \bottomrule[0.7pt]
        \end{tabular}
    }
    \vspace{-4mm}
\end{table}

\subsubsection{Different Consistency Losses}
There are three types of consistency regularization, including $\ell_1$, MSE, and soft cross-entropy consistency (adopted in the main paper), formulated as follows respectively,
\begin{small}
    \begin{align}
        \varL_{\ell_1} &= \frac{1}{|\imgseg_t|}\sum_{(i,j)\in\imgseg_t}\sum_{c=1}^C \|\pre_t^{(i,j,c)}-\pre_t'^{(i,j,c)}\|_1\,, \\
        \varL_{mse} &= \frac{1}{|\imgseg_t|}\sum_{(i,j)\in\imgseg_t}\sum_{c=1}^C \|\pre_t^{(i,j,c)}-\pre_t'^{(i,j,c)}\|_2\,, \\
        \varL_{sce} &=-\frac{1}{|\imgseg_t|}\sum_{(i,j)\in\imgseg_t}\sum_{c=1}^C \pre_t^{(i,j,c)}\log \pre_t'^{(i,j,c)}\,.
    \end{align}
\end{small}%
The result of different consistency losses adopted by \Bone-BvSB are reported in \tab~\ref{tab:consistency}.
As demonstrated, the impact of different $\varL_{cst}$ is negligible, \ie, the average mIoU changes by no more than $0.1\%$.

\subsubsection{Parameter Sensitivity}
For the parameters $\lambda_{ce}$ and $\lambda_{cst}$, \tab~\ref{tab:sensitivity} shows the sensitivity of them on ACDC under CTTA test protocol.
As $\lambda_{ce}$ and $\lambda_{cst}$ vary across a wide range, the performance only changes within a minor range (less $0.9\%$), verifying the robustness of \method framework.
For simplicity and consistency, we set  $\lambda_{ce}=1.0$ and $\lambda_{cst}=1.0$ for all experiments.

\subsubsection{Results on Cityscapes-foggy under FTTA.}
The \method framework is further evaluated on the task of Cityscapes$\to$Cityscapes-foggy.
The results are demonstrated in \tab~\ref{tab:cityfoggy}.
Consistent superior performance validates the effectiveness of \method.


\subsubsection{More Qualitative Results}
\begin{figure*}
    \centering
    \includegraphics[width=0.9\textwidth]{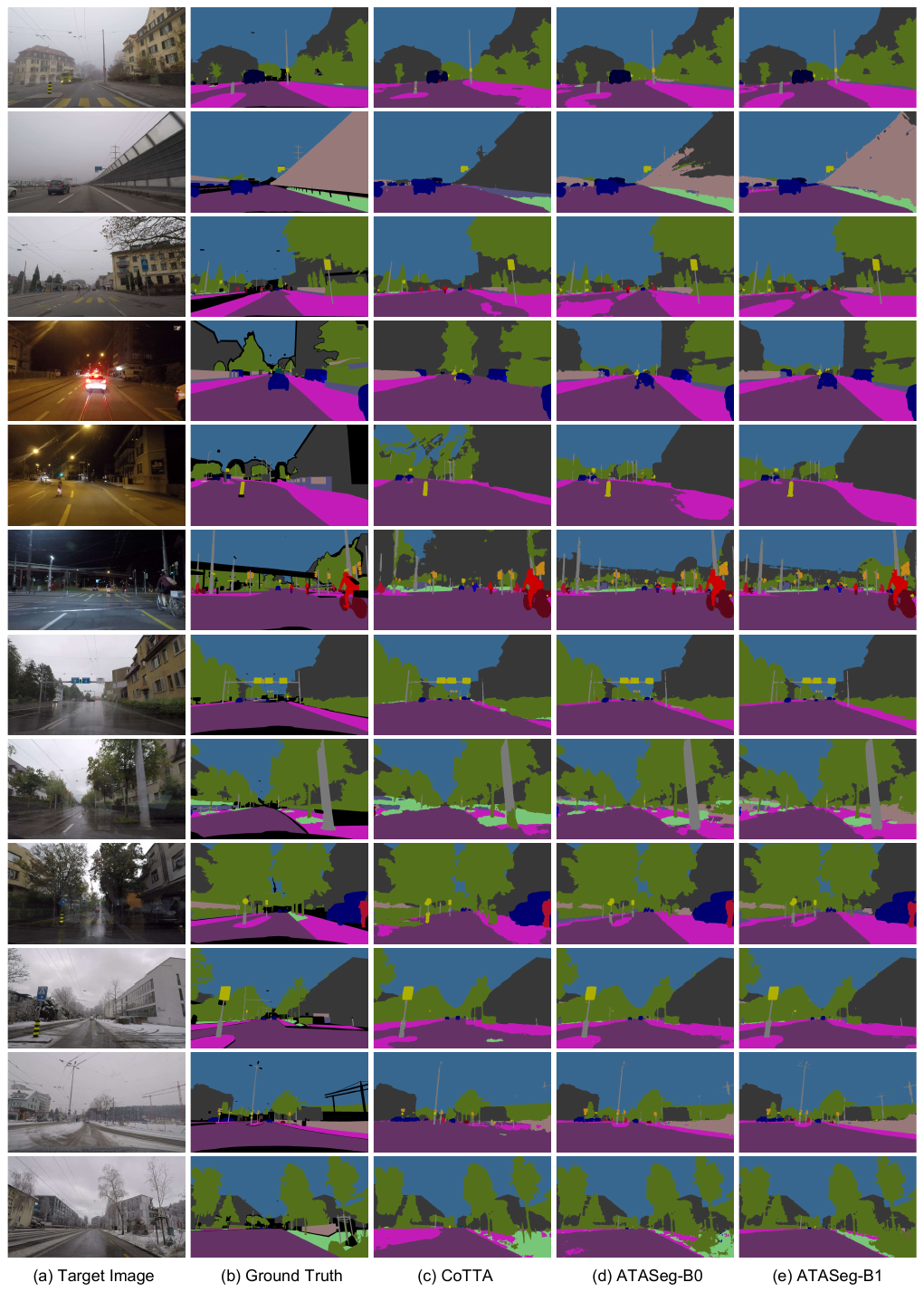}
    \caption{Visualization of {\bf segmentation results} on {\bf ACDC} under the test protocol of {\bf CTTA}. From left to right: test image, ground truth, results predicted by CoTTA, results predicted by \Bzero-BvSB and results predicted by \Bone-BvSB.}
    \label{fig:appendxi_segmap}
\end{figure*}

\begin{figure*}
    \centering
    \includegraphics[width=0.9\textwidth]{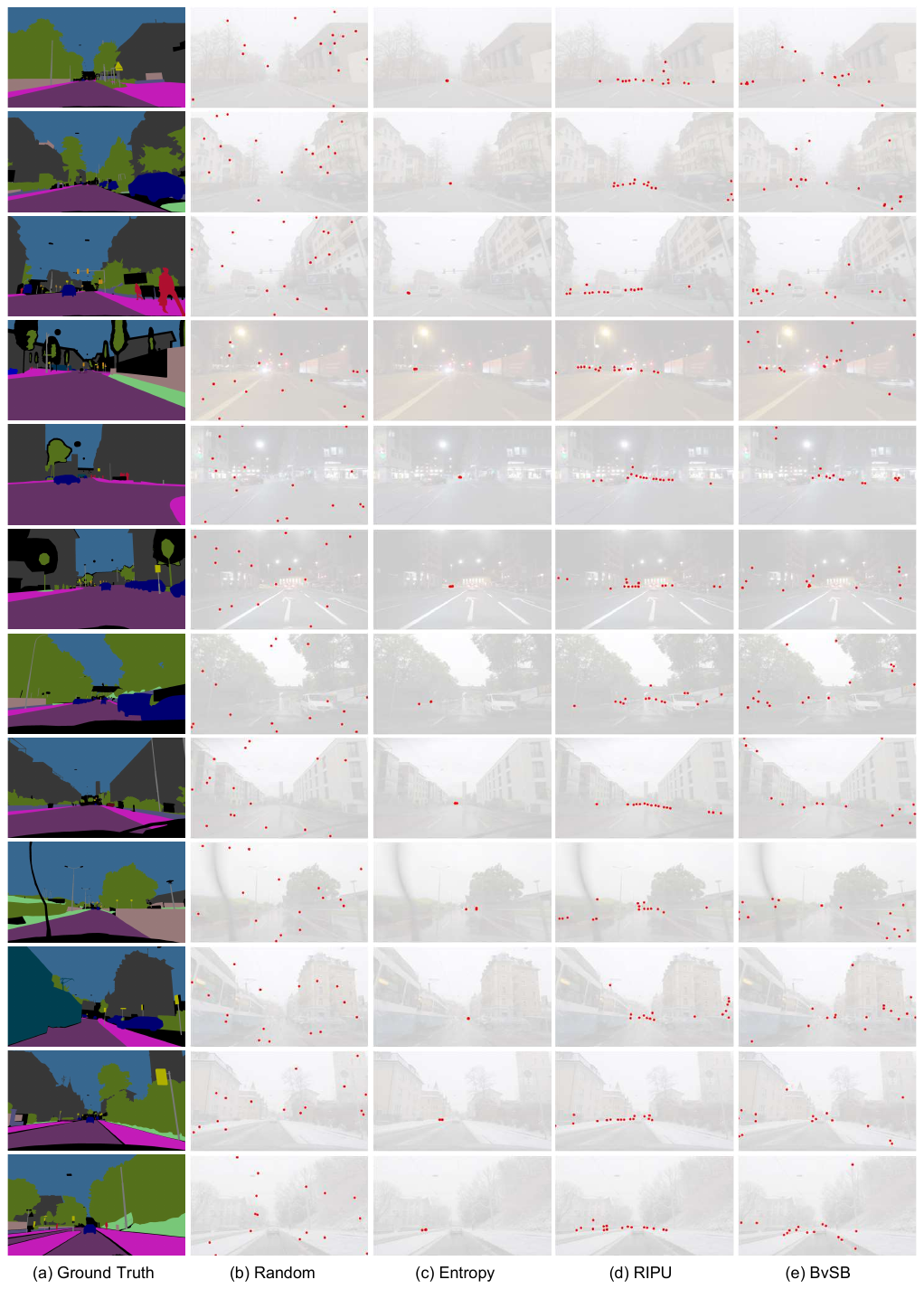}
    \caption{Visualization of {\bf selected pixels} on {\bf ACDC} under the test protocol of {\bf CTTA}. From left to right: ground truth, pixels selected by random (Rand), pixels selected by entropy (Ent), pixels selected by RIPU and pixels selected by BvSB.}
    \label{fig:appendxi_selected}
\end{figure*}

Following the main paper, this part presents more qualitative results for comparison, including segmentation maps in \fig~\ref{fig:appendxi_segmap} and selected pixels in \fig~\ref{fig:appendxi_selected}.

\subsection{Discussion}
\parastart{Societal Impact}
Our work enables a pre-trained segmentation model to adapt itself continually with very few annotation costs after deployment.
It compensates for the limited performance of previous TTA methods, which has positive impacts on communities to employ our framework in scenarios with high-security and high-accuracy requirements. 
Thus, it is economic and environmental friendliness.
We carry out experiments on public datasets and do not notice any societal issues. 
It does not involve sensitive attributes.

\parastart{Limitation}
\method achieves excellent performance on various benchmarks and test protocols as demonstrated in Section 4 in the main paper, but we still find some limitations
of it. First of all, the proposed model adapters, \ie, \Bzero and \Bone, and the label annotators, \ie, Rand, Ent, RIPU and BvSB, are preliminary attempts to perform active test-time adaptation for semantic segmentation. 
Although empirical results demonstrated the scalability of \method, better model adapters and label annotators still need to be explored to fully leverage the label information.
Even though one pixel per image achieves excellent performance, we think it may still be time-consuming to label each image.
The most ideal paradigm is to only label extremely few labels for a few images.
Finally, although we evaluate \method on various test streams, we still need to validate our approach in some real-world scenarios.

\parastart{Future Work}
Our work suggests a few promising directions for future work.
Firstly, one could improve the performance of \method by designing more advanced model adapters and label annotators.
Secondly, as with the suggested ideal paradigm, we can also develop the \method framework to be more practical by further reducing the labeling costs.
Finally, we can leverage some models to segment regions, and then label regions with a single click to further improve the efficiency of labeling.
In general, the newly proposed setup and framework have great research potential. 
We hope this work will open up new ways for adapting models in real-world scenarios with requirements of high security or accuracy.

\end{document}